\documentclass{article}

\usepackage{parskip}
\usepackage[margin=1.0in]{geometry}

\usepackage{enumitem}

\usepackage[utf8]{inputenc} 
\usepackage[T1]{fontenc}    
\usepackage{hyperref}       
\usepackage{url}            
\usepackage{booktabs}       
\usepackage{amsfonts}       
\usepackage{nicefrac}       
\usepackage{microtype}      
\usepackage{xcolor}         

\newcommand{\e}{\mathbb{E}}
\newcommand{\E}{\mathbb{E}}

\newcommand{\N}{\mathbb{N}}
\newcommand{\R}{\mathbb{R}}

\newcommand{\mcM}{\mathcal{M}}
\newcommand{\mcN}{\mathcal{N}}
\newcommand{\mcL}{\mathcal{E}}

\newcommand{\mfe}{\mathfrak{e}}
\newcommand{\mfd}{\mathfrak{d}}

\usepackage[ruled,vlined]{algorithm2e}

\usepackage{graphicx}
\usepackage{caption}
\usepackage{subcaption}
\usepackage{tikz}
\usepackage{float}

\usepackage{physics}
\usepackage{amsmath}
\usepackage{amsthm}
\usepackage{amstext}
\usepackage{amssymb}
\usepackage{thmtools}
\usepackage{siunitx}
\usepackage{bm}

\newcommand{\ceil}[1]{\lceil {#1} \rceil}

\theoremstyle{plain}
\newtheorem{theorem}{Theorem}[section]

\theoremstyle{definition}
\newtheorem{definition}[theorem]{Definition}

\theoremstyle{remark}
\newtheorem{remark}[theorem]{Remark}

\newtheorem{exmp}{Example}[section]

\usepackage[english]{babel}

\usepackage{pdfpages}

\usepackage{dsfont}
\usepackage{xcolor}
\definecolor{amber}{rgb}{1.0, 0.6, 0.0}
\usepackage{textcomp}
\usepackage{listings}
\usepackage{relsize}
\usepackage{multicol}
\usepackage{hyperref}
\hypersetup{
	colorlinks=true,
	linkcolor=black,
	filecolor=black,      
	urlcolor=black,
	citecolor=blue,
}
\usepackage[capitalize,noabbrev]{cleveref}
\crefname{figure}{Fig.}{Fig.}
\crefname{table}{Table}{Tables}
\crefname{equation}{Eqn.}{Eqns.}
\crefname{algocf}{Algorithm}{Algorithms}
\crefname{exmp}{Example}{Ex.}

\crefname{lemma}{Lemma}{Lemmas}
\crefname{corollary}{Corollary}{Cor.}
\crefname{defn}{Definition}{Definitions}

\usepackage{physics}
\usepackage{url}

\usepackage{pifont}

\usepackage{bbm}
\usepackage{mathtools}
\usepackage{hhline}
\usepackage{booktabs}
\usepackage{multirow}

\usepackage[ruled,vlined]{algorithm2e}

\title{Thinner Latent Spaces: Detecting dimension and imposing invariance through conformal autoencoders}

\usepackage{authblk}

\title{Thinner Latent Spaces: Detecting Dimension and Imposing Invariance with Conformal Autoencoders}
\author[1,2]{George A. Kevrekidis}
\author[1]{Zan Ahmad}
\author[1]{Mauro Maggioni}
\author[1]{\\Soledad Villar}
\author[1]{Yannis G. Kevrekidis}

\affil[1]{Department of Applied Mathematics and Statistics, Johns Hopkins University, Baltimore, MD, USA}
\affil[2]{Los Alamos National Laboratory, Los Alamos, NM, USA}

\date{June 23, 2025\\LA-UR-23-20785}

\begin{document}

\maketitle

\begin{abstract}
Conformal Autoencoders are a neural network architecture that imposes orthogonality conditions between the gradients of latent variables to obtain disentangled representations of data. In this work we show that orthogonality relations within the latent layer of the network can be leveraged to infer the intrinsic dimensionality of nonlinear manifold data sets (locally characterized by the dimension of their tangent space), while \textit{simultaneously} computing encoding and decoding (embedding) maps. We outline the relevant theory relying on differential geometry, and describe the corresponding gradient-descent optimization algorithm. The method is applied to several data sets and we highlight its applicability, advantages, and shortcomings. In addition, we demonstrate that the same computational technology can be used to build coordinate invariance to local group actions when defined only on a (reduced) submanifold of the embedding space.
\end{abstract}

\section{Introduction}

Dimension Reduction is a ubiquitous task in Data Science and Machine Learning. Describing apparently high-dimensional data sets using few variables when possible reduces storage, provides a better handle on the degrees of freedom of a system and their interactions; this often allows for enhanced understanding and interpretability from a human-scientific perspective, leading to more concise descriptive models.  Autoencoders \cite{kramer1991nonlinear,kingma2013auto} have broadly been used to perform dimension reduction, {\em typically requiring prior knowledge of the latent layer dimension}. In this work, we introduce an alternative computational approach, based on \textit{conformal autoencoders} \cite{parameters}, to performing nonlinear dimension reduction using autoencoder (AE) neural network (NN) architectures: our algorithm \textit{combines} the tasks of (a) inferring the dimension of a data set, and (b) computing a smooth representation map (chart), achieved with the addition of a soft orthogonality constraint (on suitable gradients of the encoding map) during training. This approach has a theoretical basis in elementary results from differential geometry. Prior knowledge of the minimal latent dimension may thus be circumvented.

In many applications, apparently high-dimensional data lie near low dimensional \textit{manifolds} (i.e. they satisfy a \textit{manifold assumption}). Dimension Reduction and Manifold Learning algorithms (including autoencoders) attempt to recover a parametrization of the underlying manifold. When the intrinsic dimension of a sampled data set is not \textit{a priori} known, we claim that requiring orthogonality of the gradients of a latent representation provides sufficient regularization for autoencoder neural networks to {\em infer} the intrinsic dimension of the data --while also computing a smooth embedding map in the process. In what follows, we give a principled account of this observation, which can be used either instead of, or in tandem with, classical nonlinear dimension reduction algorithms.

\subsection*{Background and Related Work}
For linear data sets, the tried and true linear algorithm for dimension reduction is Principal Component Analysis (PCA) and its variants \cite{jolliffe2002principal}. PCA produces eigenvector (singular vector) representations that successively explain decreasing variance in orthogonal directions. 
Given a new data point (from the original data distribution) that has not been used to generate the principal components, one may project onto the eigenvector basis to obtain a (least-squares optimal) low-dimensional representation.

For nonlinear data sets satisfying a manifold assumption, multiple state of the art constructions exist. Well known instances include Isomap \cite{tenenbaum_isomap}, Locally Linear Embedding \cite{roweis2000_LLE}, UMap \cite{McInnes2018_UMAP}, t-SNE \cite{ljpvd2008visualizing_TSNE}, Diffusion Maps  (DMaps) \cite{DiffusionPNAS} and other spectral methods, among several other examples. These methods generally rely on constructing a graph that captures geometric structure (i.e. distances) of the data set. However, due to their nonlinearity, they generally lack much of the convenience and interpretability of PCA. For instance one may no longer have a simple method to project a new data point; some form of extension of the map to new data is needed, e.g. using the Nystr\"{o}m Extension algorithm \cite{freeman2009photometric,lafon2006data,bengio2003out}. Spectral methods suffer an additional setback: the generated eigenvectors can be (and often are) {\em functionally} related: orthogonality (in Hilbert space) does not imply functional independence \cite{coifman_laffon_dmaps}. For example, even after projecting data onto DMap eigenvector components, it is not clear \textit{which} of the features are functionally independent and inferring the true (intrinsic) dimension becomes nontrivial \cite{DSILVA2018759,LMR:MGM1}. 

Nonlinear dimension reduction can also be performed using autoencoder (AE) networks. Usually, a low-dimensional bottleneck layer that separates an encoder and a decoder is used to generate the latent representation. As long as the decoder can reconstruct the AE input, no information is lost in the low-dimensional representation. 
AE objectives are often regularized in order to impose additional structure in the learned latent space. Most prominently, Variational Autoencoders (VAEs) and $\beta-$VAEs \cite{kingma2013vae, higgins2017beta}  make use of the reparametrization trick to impose statistical structure {\em within} their latent space, allowing one to \textit{sample} from a learned latent distribution. Sparse Autoencoders, which incorporate $L^1$ or $KL$-divergence penalties on network weights, encourage sparsity of the resulting network. Geometrically-inspired regularizations have been proposed in more recent work, for example $\Gamma$-VAEs \cite{kim2024gamma} use a second-order curvature-regularized objective beneficial to manifold learning on gene data sets. The closest work to us is \cite{pan2023geometric,parameters}, where several regularizations on the Jacobian of the encoder are imposed to achieve disentanglement, which is a similar objective to what we consider in the present work, albeit in a different setting. Finally, additional work \cite{rolinek2019variational,kumar2020implicit} suggests that ($\beta$-) VAEs also have implicit \textit{geometric} biases (relating to the Jacobian and Hessian of the networks) even though the regularization is a statistical one.

In general, AE architectures require \textit{a priori} knowledge of the dimension (width) of the bottle-neck layer, or at least a convenient upper bound of it, since that is hard-coded into the structure of the network before training. If the latent layer is wider than minimal, a generic autoencoder will make use of all latent nodes, producing a higher-dimensional latent embedding than necessary. However, both $\beta$-VAEs \textit{as well as} sparse autoencoders exhibit a phenomenon called mode-collapse: certain latent nodes are not `used' by the network, resulting in a lower-dimensional embedding than originally specified in the architecture. We demonstrate that \textit{conformal autoencoders} also exhibit mode collapse, but may do so more robustly due to the nature of the regularizer.

\textbf{Our contributions. }
The main contributions in this work are as follows:

\begin{itemize}
    \item[\textbf{(a)}] We develop an algorithmic framework that with a single optimization objective addresses two tasks (1) inferring the intrinsic dimension of the data {\em and} (2) computing an embedding.
    \item[\textbf{(b)}] We show we can use our algorithmic framework to compute local invariant coordinates on a submanifold of $\R^n$ when a known group action is defined on the submanifold.
    \item[\textbf{(c)}] We show that optimization objectives with geometric pointwise constraints involving NN gradients (with respect to their input) provide simple descriptors of complex global problems. The pertinent objectives can be successfully optimized using gradient descent algorithms.
\end{itemize}

\textbf{Outline. }\cref{sec:Theory_and_Methodology} outlines the relevant mathematical theory underlying our proposed numerical method. \cref{sec:Numerical_Examples} demonstrates the application of our method an illustrative synthetic data set as well as more realistic, higher-dimensional data sets arising from the solution of evolutionary Partial Differential Equations (PDEs) and  "real-world" image data. \cref{sec:symmetries_exmp} demonstrates the computation of a locally group-invariant coordinate system. 
\cref{sec:Discussion} provides a discussion. The algorithms, network architectures, additional computational examples and accompanying information can be found in the Appendices.

\section{Theory and Methodology}
\label{sec:Theory_and_Methodology}

We begin by briefly describing a differential-geometric setting for dimension reduction and its relationship to orthogonal coordinate systems. We then discuss its application specifically to autoencoders.

Througout this work, we let $\mcN\subset\R^k$ be an open, simply connected, precompact domain of `intrinsic' dimension $k$. The map $\Phi$ gives a smooth embedding of $\mcN$ into $\R^n$, $n\geq k$:
\begin{equation}
    \Phi(\mcN)\doteq\mcM\subset\R^n,
\end{equation}
and equip $\mcN$ with the metric $g$ induced by $\Phi$. In coordinate form, this is expressed in terms of the Jacobian of the embedding map, as $g=(J\Phi)^T J\Phi$. We further assume that $\mcM$ admits a single \textit{global} chart through the coordinate map $\Phi^{-1}$.

\textbf{Problem Statement:} Our primary objective is, given samples of $\mcM$, to simultaneously infer its dimension $k$ while computing a \textit{global} coordinate chart using an autoencoder architecture. 

In the usual dimension reduction setting, $n$ is the embedding dimension of a discretely-observed data set (samples of $\mcM$), and $k$ is the (low, i.e. $k\ll n$) \textit{intrinsic} dimension that is to be determined, along with a function (similar to $\Phi$) which allows for interpolation on $\mcM$. Informally, the link between the dimension of a submanifold $\mcM$ and orthogonality comes from the fact that, at any given point $p\in\mcM$, one should only be able to find at most $k$ linearly independent (and therefore also orthogonal) vectors on its tangent space $T_p\mcM$.

The assumption that a global chart exists is necessary for any autoencoder architecture as well as for other algorithms; the reader may consider our discussion as concerning \textit{a single chart} of an arbitrary manifold. While it is indeed rather restrictive geometrically, there are still interesting computational problems that satisfy that condition. There also exist recent results in the literature that discuss extending single chart methods to atlases (e.g. \cite{schonsheck2020chart}), as well as techniques that find provably spatially extended charts and can be readily extended to atlases \cite{jms:UniformizationEigenfunctions,jms:UniformizationEigenfunctions2}; see also \cite{georgiou2023locating,bello2023gentlest}. The additional requirement of $\Phi$ being conformal, is more restrictive and less studied computationally. We comment on it throughout the text. The mathematical background of our work is shared with the methods developed in \cite{lin2013parallel}. There, vector fields are \textit{first} generated on estimated tangent spaces and \textit{subsequently} integrated to obtain coordinates. However, the dimension inference step is reduced to estimation of the tangent space. Our method, instead uses the neural network gradients to generate vector fields which are integrable by definition, and subsequently optimizes network weights to satisfy constraints expressed through conditions on the vector fields.

\subsection{Formal Statement}
\label{sec:Formal_Statement}
In an idealized setting, we have the following intuitive result:

\begin{theorem}[Orthogonal Charts]
	Let $\mcN,\Phi,\mcM$ be defined as above and $f:\R^n\to\R^n$ be a smooth function such that its restriction $f\vert_\mcM:\mcM\to\R^n$ is smoothly invertible on its image. Assume $f$ satisfies
	\begin{equation}
		\mcL f:=\sum_{\stackrel{i=1}{j>i}}^n\abs{\expval{\grad_p^\mcM f_i,\grad_p^\mcM f_j}}^2=0\qc\forall p\in\mcM
		\label{eqn:orthogonal_charts}
	\end{equation}
where each component $f_i\in C^\infty(\mcM,\R)$, $\grad_{p}^\mcM f_i$ is the orthogonal projection of $\grad_{p}{f_i}$ onto $T_p\mcM$ and $\expval{\cdot,\cdot}$ is the $\ell^2$ inner product on $\R^n$. Then $f\vert_\mcM$ has exactly $k$ non-constant functionally independent components $f_{i_1},\dots,f_{i_k}$, and its restriction to these components, $f\vert_k:\mcM\to\R^k$, with $f\vert_k(x):=(f_{i_1}(x),\dots,f_{i_k}(x))$, is a smooth chart for $\mcM$.
\label{thm:orthogonal_charts}
\end{theorem}

For a proof, see \cref{app:Diff_Geo}.
\textit{If} a function $f$ that satisfies the conditions outlined in \cref{thm:orthogonal_charts} exists, it determines both the intrinsic dimension \(k\) and a smooth chart $(\mcM,f\vert_k)$. A sufficient condition for the existence of such $f$ is that $\mcM$ is \textit{conformally flat}, i.e. its Weyl tensor vanishes \cite{DeTurckDennisM.1984Eoed}. However, such regular coordinate maps rarely do exist for arbitrary submanifolds $\mcM$, which generically do not admit a coordinate system in which the metric has diagonal form even if they admit a global chart. We give a more thorough account of this condition in \cref{app:Diff_Geo} while an extensive account in the context of dimension reduction can be found in \cite{lin2013parallel}.

One may consider a less restrictive condition by replacing $\grad_p^\mcM$ with $\grad_p$:
	\begin{equation}
		\mcL f=\sum_{\stackrel{i=1}{j>i}}^n\abs{\expval{\grad_p f_i,\grad_p f_j}}^2=0\qc\forall p\in\mcM
		\label{eqn:relaxed_orthogonal_charts}
	\end{equation}
which is satisfied as long as there exists some conformally flat submanifold that is nowhere normal to an embedded submanifold (e.g. a hyperplane). This condition can also be satisfied by functions with `wrong' latent dimension however, such as any canonical coordinatization of the embedding space $\R^n$. Nevertheless, both \cref{eqn:orthogonal_charts} and \cref{eqn:relaxed_orthogonal_charts} can act as useful regularization constraints that can be imposed directly in the latent space of autoencoder architectures.

\subsection{Numerical Methods}
\label{sec:Numerical_Method}
We will use a conformal autoencoder (CAE) architecture to learn a prescribed orthogonal latent space. 

We define the architecture as follows:

\begin{definition}[Conformal Autoencoder (CAE)] An \textit{autoencoder} (AE) consists of a pair of feed-forward networks (\cref{definition:FFNN}), an encoder $\mfe$ and a decoder $\mfd$, whose weights are optimized such that $\mfe\circ\mfd=i$ is the identity map, i.e. the decoder is the encoder's right inverse. Of particular importance is its latent layer representation: the components of $\mfe$ (resp. input of $\mfd$) which we denote by $\bm{\nu}(\vb{x})=(\nu_1,...,\nu_l)(\vb{x})=\mfe(\vb{x})\in\R^l$, where $l$ is a positive integer. A \textit{conformal autoencoder} (CAE) is an autoencoder whose latent layer satisfies additional conditions of the form
	\begin{equation}
		\expval{\grad\nu_i({\vb{x}}),\grad\nu_j({\vb{x}})}=0
		\label{eqn:CAE_ortho_constraints}
	\end{equation}
for all inputs $\vb{x}$ and for $\qty{i,j\in[l]:i\neq j}$; here $\expval{\cdot,\cdot}$ is a pre-specified inner product. Note that this is equivalent to the Jacobian of the encoder being diagonal.
\end{definition}

Below we describe three variants of our proposed algorithmic framework that uses CAEs and \cref{thm:orthogonal_charts} to simultaneously learn the dimension of the data and the embedding (see \cref{app:Algs} for a complete description).

\textbf{\cref{alg:CAE}.}
We assume that we are given a set of $N$ discrete observations of the form
\begin{equation}
	\qty{\vb{x}_i}_{i=1}^N = \qty{(x_1,...,x_n)_i}_{i=1}^N
\end{equation}
with each $\vb{x}_i\in\mcM\subset\R^n$, where $\mcM$ is an unknown precompact submanifold of $\R^n$ of unknown dimension $k$ that admits a single chart. The goal is to simultaneously learn the dimension of the manifold and an smooth embedding using an approximation of \cref{thm:orthogonal_charts}.

For an encoder $\mfe$, we let $\bm{\nu}_i:=\mfe(\vb{x}_i)$.
We define an encoder-decoder pair $(\mfe,\mfd)$ with \textit{a ``full"} $n$-dimensional latent space, and consider the following loss function (based on \cref{eqn:relaxed_orthogonal_charts}):

\begin{equation}
    \mcL_\text{CAE}=\underbrace{\frac{1}{N}\sum_{i=1}^N\norm{\vb{x}_i-\hat{\vb{x}}_i}_2^2}_{\text{reconstruction term}}\\+\underbrace{\alpha\frac{1}{N}\sum_{i=1}^N\sum_{j>k}\abs{\expval{\grad\nu_j({\vb{x}_i}),\grad\nu_k({\vb{x}_j})}}^2}_{\text{orthogonality term}}\label{eqn:CAE_loss}
\end{equation}
    
\noindent where $\hat{\vb{x}}_i=\mfd\circ\mfe(\vb{x}_i)$, $\alpha>0$ is a positive constant, and $\expval{\cdot,\cdot}$ is the Euclidean inner product on $\R^n$. The reconstruction term ensures the invertibility of the encoder on its range. \cref{alg:CAE}, described in Appendix \ref{app:Algs}, imposes both orthogonality and invertibility as a soft constraints by minimizing \cref{eqn:CAE_loss}.

\begin{remark} \label{rem:issue} 
If the autoencoder satisfies \cref{thm:orthogonal_charts} (with $f=\mfe$) then $\mcL_\text{CAE}=0$ on the entirety of $\mcM$ for $\bm{\nu}=\mfe(\vb{x})$. However, the converse does not hold, since generically $\grad^\mcM\neq\grad$ and an orthogonal projection onto a linear subspace (such as the tangent space to $\mcM$ at a point) does not preserve orthogonality between vectors. Effectively, in order to have \cref{thm:orthogonal_charts} as a \textit{guarantee} of having inferred the correct, minimal dimension, we would like equation \cref{eqn:CAE_loss} to make use of an estimate of $\grad^\mcM$ when computing the orthogonality term, thus approximating \cref{eqn:orthogonal_charts}. This information is not directly available during optimization, since it requires knowledge of the embedding map $\Phi$ which is still unknown. 
\end{remark}

\textbf{\cref{alg:CAE_proj}.} 
If the intrinsic dimension $k$ is known and the goal is to compute an orthogonal chart, one may pre-compute local tangent spaces {\em at each point} of the given data. This is what \cref{alg:CAE_proj} does. It performs principal component analysis (PCA) each points' $K$-nearest neighbors (as done in \cite{lin2013parallel}), or in a principled multiscale fashion for multiple values of $K$ \cite{LMR:MGM1,CM:MGM2}. This process would yield a good estimate for $\grad^\mcM$, as well as estimate the dimension of the tangent space. For details, see \ref{app:Algs}).

\textbf{\cref{alg:CAE_orthog}.} Furthermore, provided a model that already performs the dimension reduction task without providing an orthogonal chart, we may still want to orthogonalize the gradients of the inferred coordinates on $T\mcM$. This is relevant, for example, in the invariance setting of \cref{sec:symmetries_exmp}. It can be achieved with a very similar algorithm, as proposed in \cref{alg:CAE_orthog}, as long as gradients of the original model are computable.

\begin{remark}
    The mathematical background described above is shared with the methods developed in \cite{lin2013parallel}. There, vector fields are \textit{first} generated on estimated tangent spaces and \textit{subsequently} integrated to obtain coordinates. However, the dimension inference step is reduced to estimation of the tangent space. Our method, instead uses the neural network gradients to generate vector fields which are integrable by definition, and subsequently optimizes network weights to satisfy constraints expressed through conditions on the vector fields.
\end{remark}

\section{Numerical Examples}
\label{sec:Numerical_Examples}
The numerical examples of this section showcase the behavior of the proposed dimension reduction algorithm (\cref{alg:CAE}). The first toy example is included for pedagogical purposes. Higher-dimensional PDE examples demonstrate its applicability in an exploratory setting in which it may be more challenging to implement known dimension reduction techniques, and the bobblehead example demonstrates the approach for a high-dimensional image data set which has intrinsic (nontrivial) geometric structure. Finally, we provide \textit{some} benchmarking analysis on these examples comparing to VAE and Sparse autoencoder architectures. Additional numerical examples are provided in \cref{app:numerical_examples}, along with expanded benchmarking results.

\subsection{Synthetic Examples}
\label{sec:synthetic_examples}
\begin{exmp}[Toy]
\label{exmp:toy}

Let $x,y$ be the canonical coordinates in $\R^2$ and define $\Phi:\R^2\to\R^3$ as
\begin{align}
\label{eqn:exmp_toy}
\Phi(x,y)=\mqty(4x\sin(y)\\xy^2\\20\frac{\cos(x)}{y^2+1})
\end{align}
We consider the image of the square $[1,2]^2$ under $\Phi$, which is a two-dimensional manifold embedded in $\R^3$. We normalize the components of $\Phi$ to lie within the 3-dimensional cube, and sample $N=2500$ points uniformly at random from $\Phi([1,2]^2)$ with Gaussian ambient noise ($\sigma=0.1$). We train a conformal autoencoder on this data using the loss from \cref{eqn:CAE_loss} with a three-dimensional latent layer $\nu\in\R^3$.

In \cref{fig:ex_1_training} we present the result of applying the proposed algorithm to this synthetic data set \cref{eqn:exmp_toy}. By plotting $\E_{\vb{x}}\norm{\grad \nu_i(\vb{x})}_2$ over the data as a function of the training epoch (\cref{fig:ex_1_grads}), we see how the network `searches' across dimensions in its effort to both fit the data and reduce the dimension: A positive value corresponds to a component being used. Importantly, one of the components (\textcolor{red}{red line}) collapses to zero, since the network is capable of minimizing the loss function by only making use of 2 dimensions eventually (the blue and yelow components). In \cref{fig:ex_1_vals} we demonstrate that, indeed,  the first (\textcolor{blue}{blue}) and third (\textcolor{amber}{yellow}) components vary across the training points, while the second (\textcolor{red}{red}) is a constant, and hence uninformative about the manifold from the autoencoder's perspective.

\begin{figure}[h]
	\begin{subfigure}[h]{0.37\textwidth}
		\includegraphics[width=\textwidth]{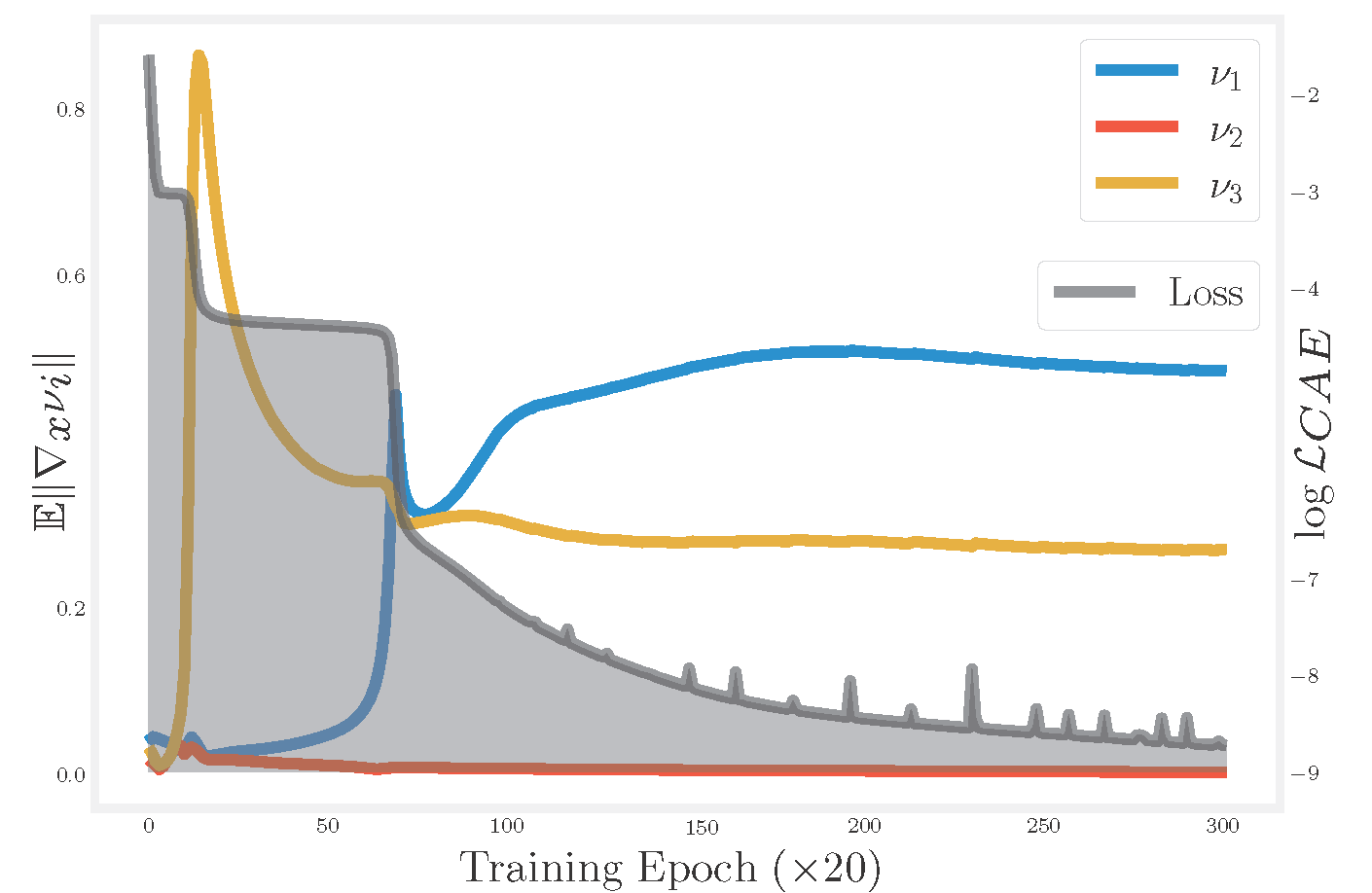}
		\caption{}
		\label{fig:ex_1_grads}
	\end{subfigure}
	\begin{subfigure}[h]{0.37\textwidth}\includegraphics[width=\textwidth]{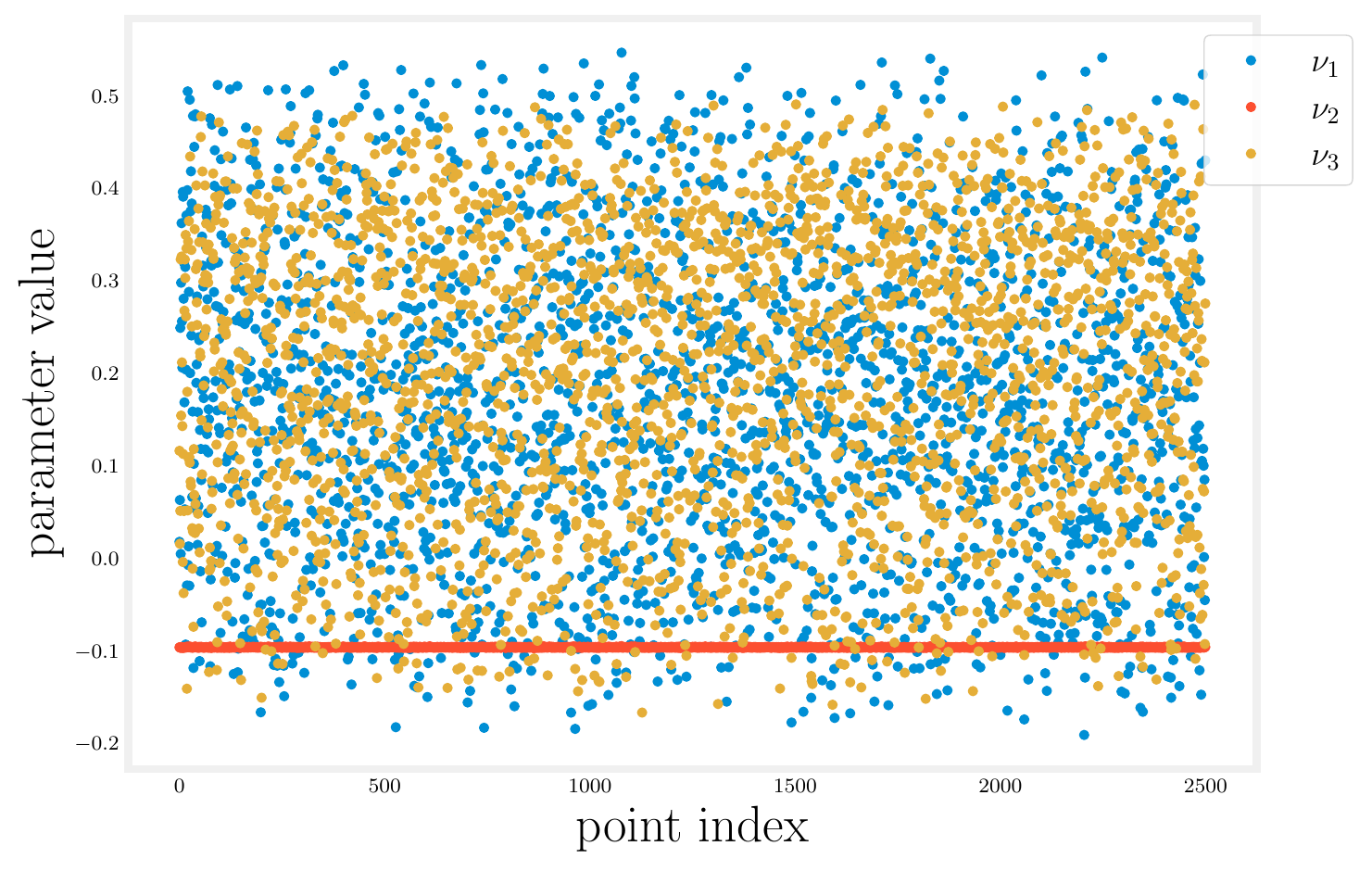}
		\caption{}
		\label{fig:ex_1_vals}
        \end{subfigure}
        \begin{subfigure}[h]{0.25\textwidth}
            \includegraphics[width=\textwidth]{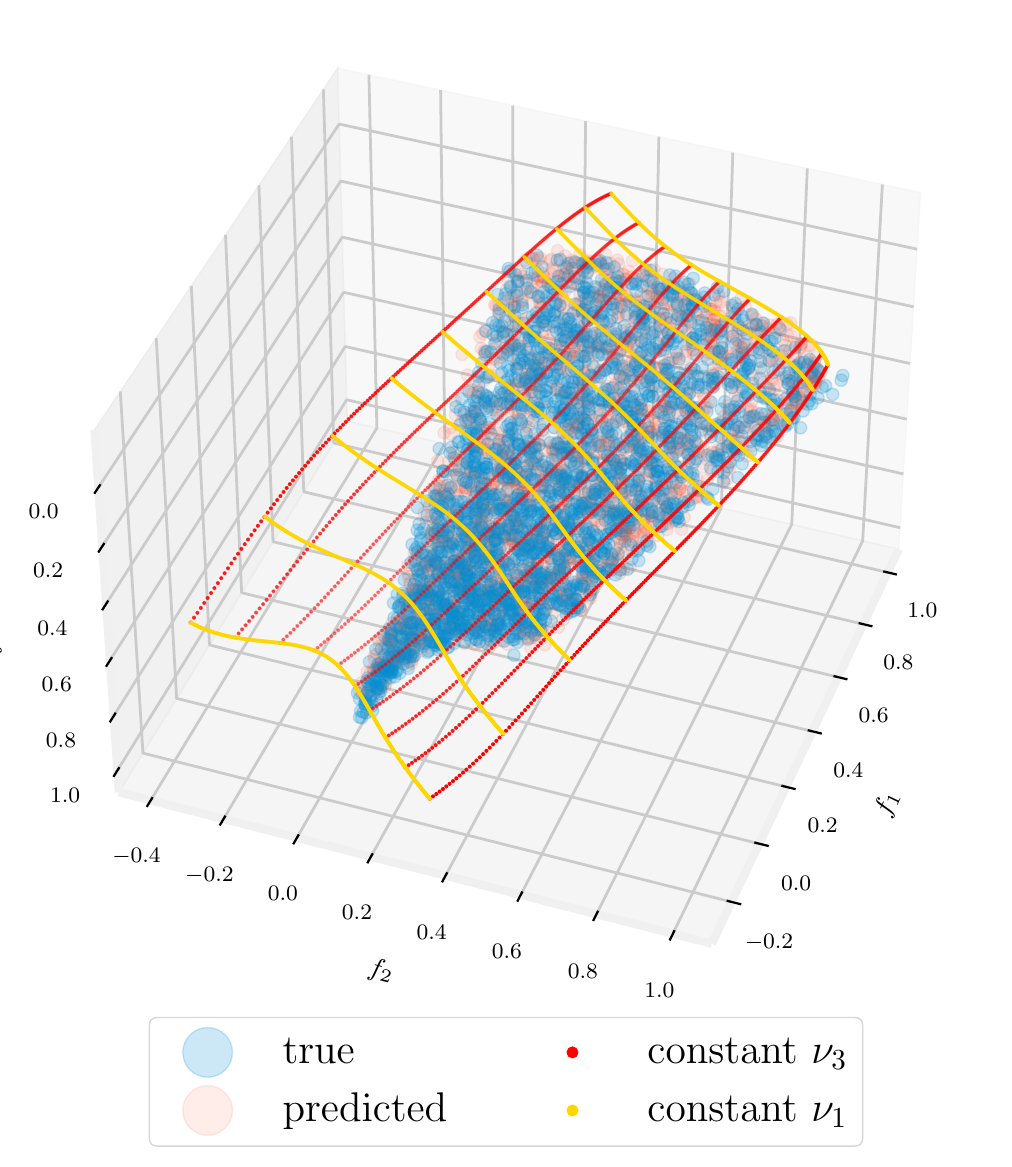}
            \caption{}
            \label{fig:ex_1_charts}
        \end{subfigure}
	\caption{Optimization result for a single run of \cref{alg:CAE} on the data set described in \cref{exmp:toy}. \cref{fig:ex_1_grads} depicts the expected value of the norm gradients over the training points varying during optimization (left axis), along with the corresponding decreasing loss, $\mathcal{E}_\text{CAE}$  (right axis). \cref{fig:ex_1_vals} shows the resulting values of the latent variables for each training point after convergence, demonstrating the collapse of $\grad\nu_2$ over the data submanifold, on which $\nu_2$ is constant. Note the early excursion of the estimated latent dimension during training up to a high of three (around training epoch 10) before collapsing back to two. \cref{fig:ex_1_charts} plots the ground truth (\textcolor{blue}{blue}) and predicted (\textcolor{orange}{orange}) manifold samples, along with level sets of the predicted two-dimensional chart (\textcolor{red}{constant $\nu_3$}, \textcolor{amber}{constant $\nu_1$}). The two plots show different perspectives of the same object.}
    \label{fig:ex_1_training}
\end{figure}

In \cref{fig:ex_1_charts} we visualize the level sets of the obtained two-dimensional chart on the manifold. The chart is reconstructed by setting $\nu_1$ equal to its mean over the training set (from \cref{fig:ex_1_vals}, we notice that $\nu_1$ is \textit{practically} constant over the data, so we substitute it by its mean). The fact that the two-dimensional map spans the manifold confirms that the additional direction is uninformative, and that the decoder truly offers a 2-dimensional approximation of the $\R^3$-embedded data set (recall that due to the presence of some noise, this will not be an exact map). To verify, as \cref{fig:ex_1_training} suggests, that the autoencoder has truly found a two-dimensional representation of the surface, we generate and encode an additional sample of $N=10,000$ points of the surface in $\R^3$, and encode it using the trained network. We then set the constant latent variable to be equal to its mean over the training set for each test point $(\nu_2^\text{test}=\e[\nu_2^\text{train}])$, and use the decoder to reconstruct the (now exactly) two-dimensional sample in 3-D, before computing the $L^2$ test error between the input and reconstruction. This yields training error $\mcL_\text{CAE}=$\num{1.6e-4} and test error $L^2_\text{test}=$\num{1.5e-4}. Note that we did not incorporate the  orthogonality loss component in the test error calculations. It is informative to consider the behavior of the algorithm given different ambient dimension and level of noise on the intrinsically low-dimensional data. In \cref{ap:robustness} we give a computational account for the robustness of the proposed algorithm, centered around this particular synthetic data set.

\end{exmp}

\subsection{PDE examples}\label{exmp:PDE}

A typical example of model reduction in the case of dissipative PDEs arises when those are known (or suspected) to possess an {\em inertial manifold}: a finite-dimensional, smooth, attracting invariant manifold that contains the global attractor (the long-term PDE dynamics) and attracts all solutions exponentially quickly \cite{Foias1, Foias2}. 
The theory of inertial manifolds, and the theory and algorithms of numerically approximating them, were developed in the late 1980-early 1990 years \cite{Yannis1,Yannis2}; machine learning tools and algorithms are currently causing a renewed interest in this research direction (\cite{koronaki2023nonlinear,linot2022data,anirudh2020improved,lee2020model}). In principle, instead of parametrizing the (approximate) inertial manifold in terms of the low order eigenfunction of a linearized version of the problem operator, a data-driven parametrization can be obtained using an autoencoder \cite{kramer1991nonlinear,kingma2013auto,koronaki2023nonlinear}.

We apply \cref{alg:CAE} on two ``high-dimensional" data sets, arising from a spectral discretization of two model dissipative PDEs known to possess an Inertial Manifold:
the Kuramoto-Sivashinsky (KS) PDE and the Chaffee-Infante (CI) PDE. The data have been obtained from regularly sampling time series from (empirically converged) spectral discretizations of the PDEs, one using eight Fourier modes (KS) and one using ten Fourier modes (CI). It is known (\cite{Yannis1,Yannis2,Foias2}) that the inertial manifold (and thus, the minimal latent space) is three-dimensional (KS) and two-dimensional (CI) respectively, for the corresponding parameter values we use.  The data sets have been rescaled to have features in the unit cube $([0,1]^n)$. In these examples we set apart a percentage of the given data sets to use for testing after optimizing the network.

\begin{exmp}[KS]\label{exmp:KS}
The full data set consists of $N=1857$ data points embedded in $\R^8$. We use \cref{alg:CAE} on $50\%$ of the data set. The evolution of the CAE training  is shown in \cref{fig:ex_4_KS_gradient_loss_main_text}. The final training error is $\mcL_\text{CAE}=\num{2.9e-4}$ and the test error, computed on the test set after setting all redundant latent parameters (except $\nu_3$, $\nu_6$, $\nu_7$) equal to their means during training, is $\mcL=\num{3.0e-4}$. We note again (in the -color coded- trajectories of the latent component gradients) that several components ``become active" simultaneously during training, only for some of them to ``collapse back" later on.  \cref{fig:PDE_embeddings_KS_main_text} illustrates the final 3-dimensional data-driven embedding, colored by the first ambient component of the data $x_2$, which can be seen to vary smoothly on the embedding.

\begin{figure}[h!]
\centering
	\begin{subfigure}[b]{0.45\textwidth}\includegraphics[width=\textwidth]{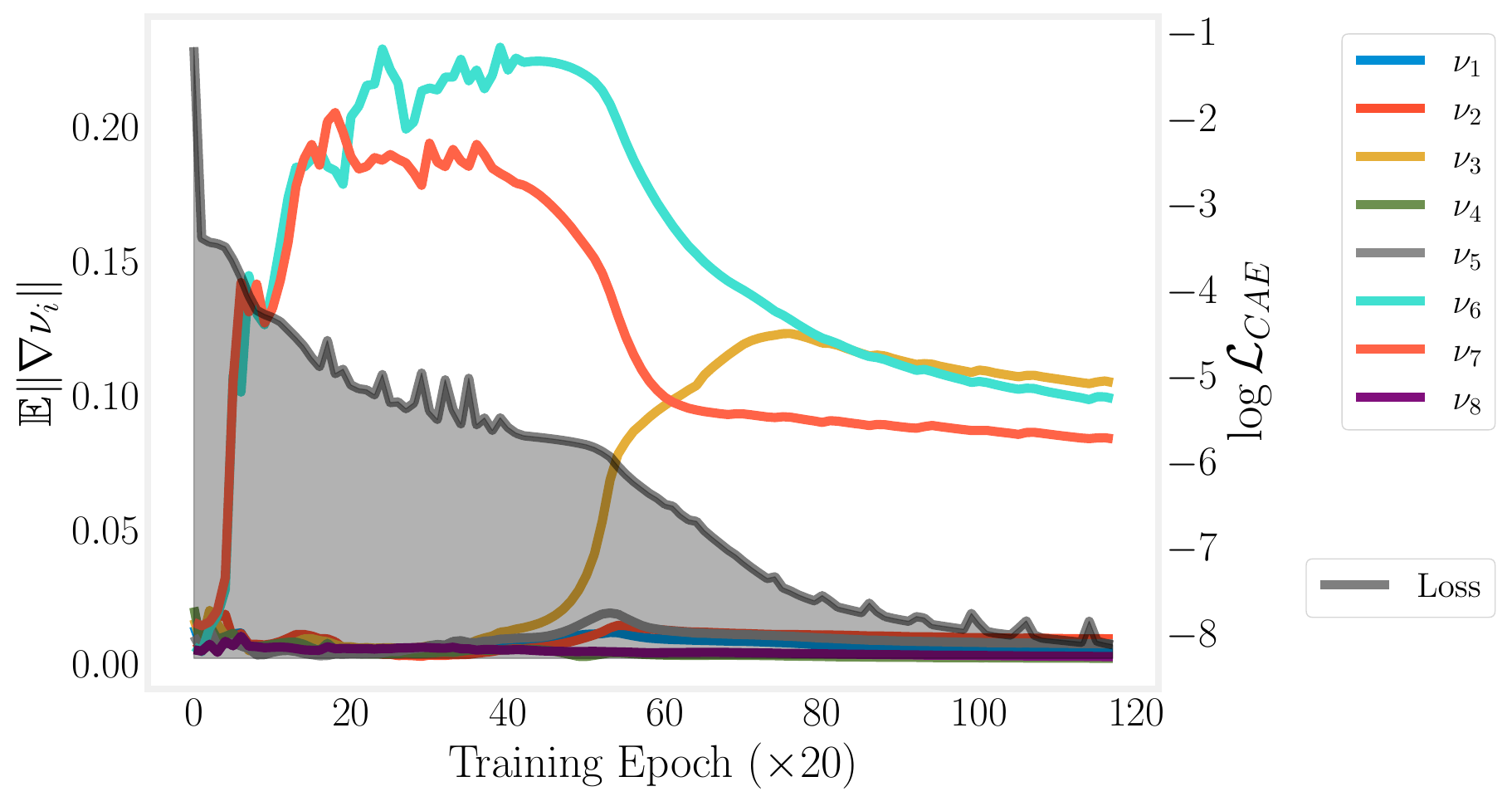}
		\caption{}
		\label{fig:ex_4_KS_gradient_loss_main_text}
	\end{subfigure}
        \begin{subfigure}[b]{0.30\textwidth}
		\includegraphics[width=\textwidth]{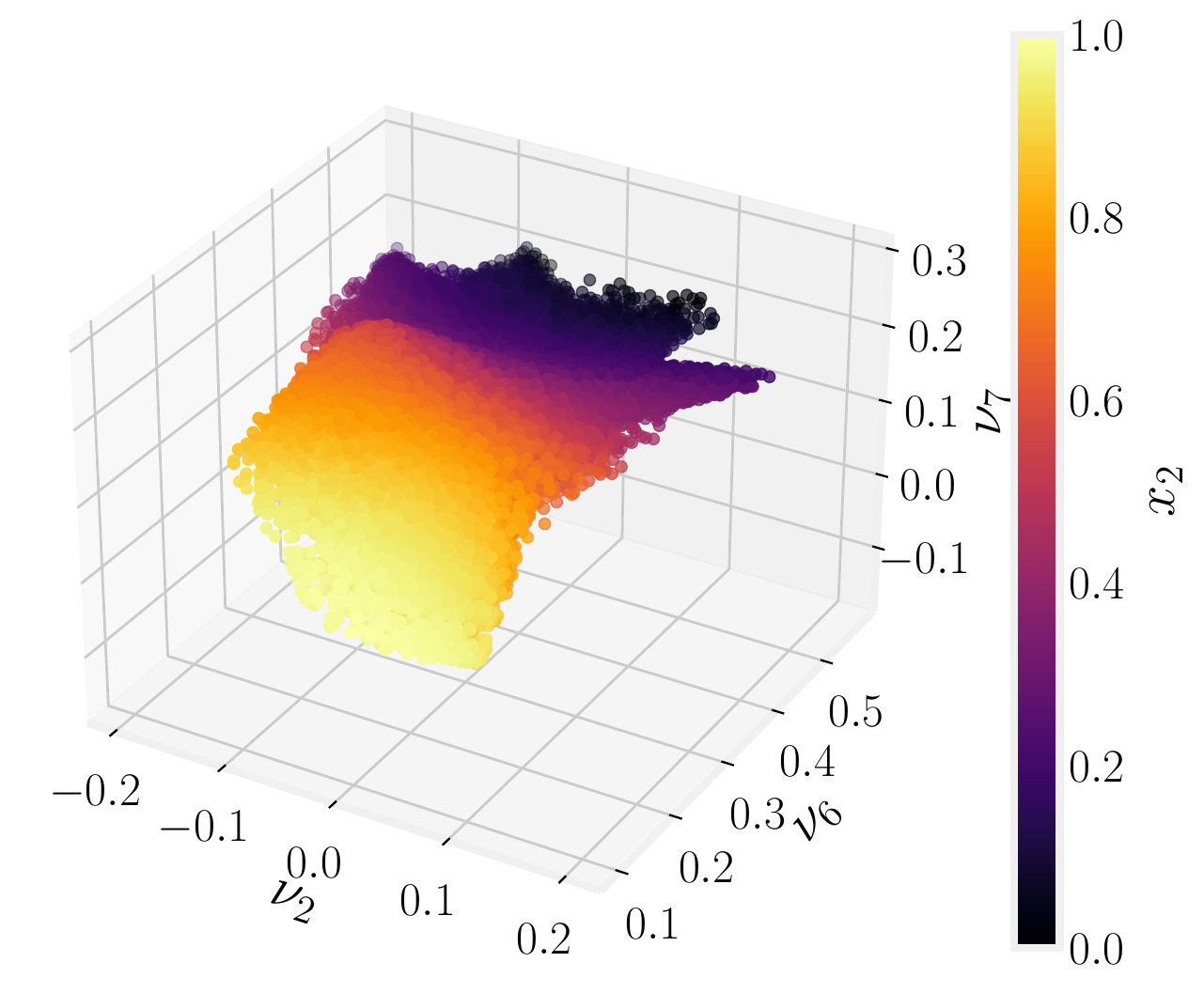}
		\caption{}
		\label{fig:PDE_embeddings_KS_main_text}
	\end{subfigure}
	\caption{(a) Training trajectory of \cref{alg:CAE} on the KS data set. (b) Data-driven embedding from \cref{exmp:KS} colored by one of the ambient coordinates of the given data set. }
	\label{fig:KS_main_text}
\end{figure}

\end{exmp}

\begin{exmp}[CI]\label{exmp:CI}
	The second full data set consists of $N=3606$ points in $\R^{10}$. We use \cref{alg:CAE} on $80\%$ of the data set. The training trajectory is represented in \cref{fig:ex_4_CI_gradient_loss} along with a PCA fit on the full data set for reference. The final training error is $\mcL_\text{CAE}=\num{6.7e-4}$, and the test error, computed on the test set after setting all redundant latent parameters (except $\nu_3$, $\nu_{10}$) equal to their training means, is $\mcL=\num{7.0e-4}$. Notice once again the intermittent activation of the latent component gradients. \cref{fig:PDE_embeddings_CI} depicts the final two-dimensional data-driven embedding, colored by the first ambient component of the data $x_1$, which can be seen to vary smoothly along the embedding. In \cref{fig:ex_4_PCA} (Appendix) we include a PCA explained variance fit for some intuition on how a low-dimensional linear approximation would perform on the specific data sets.
\end{exmp}

\subsection{Bobbleheads}
\begin{figure}
    \centering
    \begin{subfigure}[b]{0.30\textwidth}
		\includegraphics[width=\textwidth]{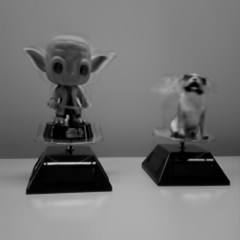}
		\caption{}
		\label{fig:bobblehead_input}
    \end{subfigure}
    \begin{subfigure}[b]{0.30\textwidth}
		\includegraphics[width=\textwidth]{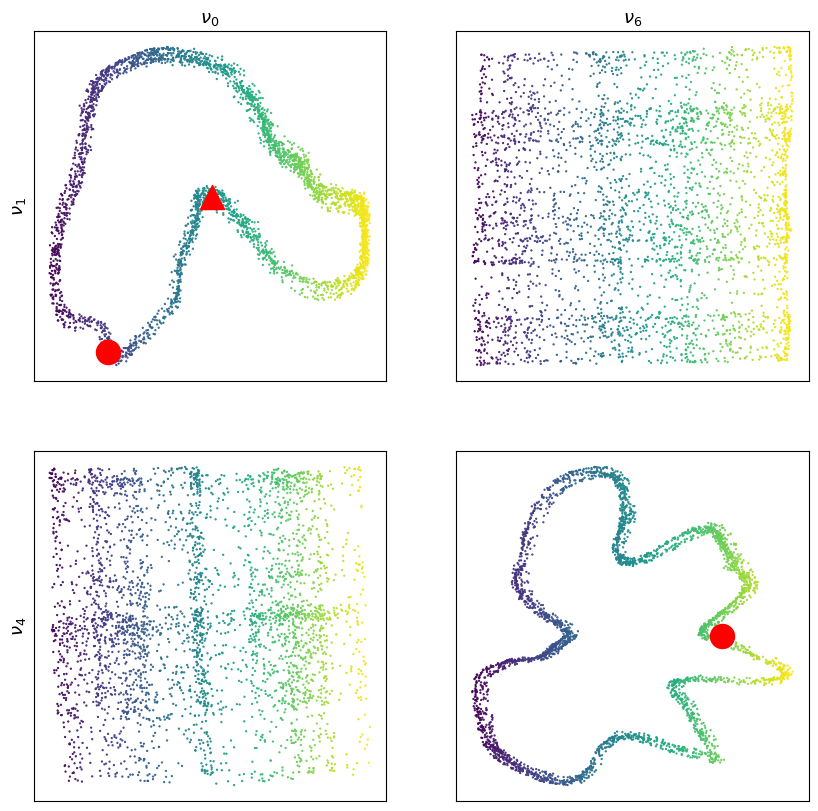}
		\caption{}
		\label{fig:bobblehead_latent}
	\end{subfigure}
    \begin{subfigure}[b]{0.30\textwidth}
		\includegraphics[width=\textwidth]{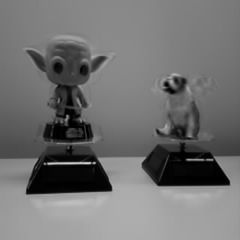}
		\caption{}
		\label{fig:bobblehead_generated}
    \end{subfigure}
    \caption{\textbf{(a)} Reconstructed training image of the two bobbleheads \textbf{(b)} Latent Representation of the discovered four-dimensional latent space (which parametrizes a 2-D Torus in 4-D) The embedding of subfigure (a) is denoted by red circles  \textbf{(c)} Generated image recovered from varying parameters of the top left latent component, denoted by a red triangle. We see that these components correspond to a rotation of the bulldog.}
    \label{fig:bobbleheads}
\end{figure}

The bobblehead data set was introduced in \cite{lederman2014common}. We consider only part of the original data set, which consists of \textit{one} camera taking snapshots of \textit{two} figurines that are rotating at different, unknown fixed frequencies ($\omega_1,\omega_2$) as in \cref{fig:bobblehead_input}. Even though the images themselves are high-dimensional (pixels), the position of each figure may be parametrized by an \textit{angle}, thus satisfying the manifold assumption and giving us a map $\Phi$ between image space and the 2-dimensional torus T2 (embedded in 4-D). To recover this latent structure, we first project the images onto their first $60$ principal components\footnote{Here, one may make a different choice such as a random, Fourier, or wavelet projection.} and use a CAE with an overspecified 8-dimensional latent layer. After succesful training of the network, we observe that only four latent components are used, and these recover a parametrization of the 2-D Torus (\cref{fig:bobblehead_latent}), i.e the cross product of two (topological) circles (see also \cite{scoccola2022toroidal}). After obtaining this latent space with a small CAE network, we train a larger decoder network to more-accurately predict the truncated PCA components of the images, and show that we can generate images that were not necessarily in the training set by varying corresponding parameters in the latent space (\cref{fig:bobblehead_generated})

While a PCA projection is not necessarily robust for image data sets (often producing artifacts), we use this example to showcase the dimension reduction capability of the proposed algorithm. Modifications with respect to the architecture could possibly yield better results on image data. We describe the technical details in \cref{ap:bobbleheads}.

\subsection{Benchmarking Autoencoder Variants}

We benchmark the Conformal Autoencoder (CAE) against several autoencoder variants on the CI and KS datasets of \cref{sec:Numerical_Examples}. These include: a standard fully-connected autoencoder without regularization (Simple AE); a sparse autoencoder with an $L^1$ penalty on latent activations weighted by $\alpha$; a sparse autoencoder with a Kullback-Leibler (KL) divergence penalty encouraging the average activation of each latent unit to match a target sparsity $\rho$ (weighted by $\alpha$); an autoencoder penalizing the squared gradient norms $\|\nabla_x \nu_i\|^2$ of latent variables $\nu_i$ with respect to input features (Gradient Norm AE); and a $\beta$-VAE (standard Gaussian prior) that scales the KL divergence term by a fixed weight $\beta$. For each model, we sweep over the regularization parameters $\alpha$, $\beta$, and $\rho$, and report the configuration yielding both low reconstruction error and the most accurate estimate of the intrinsic dimension (summarized in \cref{tab:ae-benchmark}). The intrinsic dimension is estimated post-training by progressively masking low-variance latent coordinates (setting them to their dataset mean) and measuring the impact on reconstruction fidelity; the smallest number of latent variables needed to preserve reconstruction quality is reported as the dimension. Details regarding model architecture, training setup, and additional results for different parameter values appear in Appendix~\ref{app:benchmark_specs}. Overall, we observe that CAEs and $L^1$-regularized sparse AEs perform well in simultaneously reducing dimension and reconstructing the data. Other methods to exhibit mode collapse, but do so in the expense of reconstruction error.

\begin{table}[h]
\centering
\caption{Estimated intrinsic dimension and reconstruction error $\mathcal{E}$ across autoencoder models trained on the Chafee-Infante (CI) and Kuramoto-Sivashinsky (KS) PDE datasets.}
\label{tab:ae-benchmark}
\begin{tabular}{lcccccc}
\toprule
\textbf{Model} & \textbf{Param(s)} & \multicolumn{2}{c}{\textbf{CI PDE}} & \multicolumn{2}{c}{\textbf{KS PDE}} \\
\cmidrule(lr){3-4} \cmidrule(lr){5-6}
& & \textbf{ID} & $\mathcal{E}$ & \textbf{ID} & $\mathcal{E}$ \\
\midrule
CAE (ours) & $\alpha = 1.0$ & \textbf{2} & $3.6 \times 10^{-5}$ & \textbf{3} & $4.4 \times 10^{-5}$ \\
Simple AE & -- & 10 & $4.8 \times 10^{-5}$ & 8 & $3.5 \times 10^{-5}$ \\
Sparse AE ($L^1$) & $\alpha = 0.1$ & 2 & $5.7 \times 10^{-4}$ & 3 & $4.2 \times 10^{-4}$ \\
Sparse AE (KL) & $\rho = 0.01$, $\alpha = 0.01$ & 8 & $2.4 \times 10^{-4}$ & 5 & $3.0 \times 10^{-4}$ \\
Gradient Norm AE & $\alpha = 0.01$ & 8 & $2.3 \times 10^{-4}$ & 8 & $5.1 \times 10^{-4}$ \\
Beta-VAE & $\beta = 10^{-5}, 10^{-4}$ & 10, 10 & $3e\text{-}5$, $1.3e\text{-}4$ & 8, 8 & $3.5e\text{-}5$, $1.7e\text{-}4$ \\
\bottomrule
\end{tabular}
\end{table}

\section{Symmetries and Invariants}

Orthogonality conditions can be used to describe (local) invariance with respect to smooth group actions by identifying coordinates that are invariant with respect to the group action and local sections which contain one point per orbit. Informally, if $X\in\mathfrak{g}$ is an element of the Lie algebra of the group $G$ acting freely on the embedded submanifold $\mcM\subset\R^n$, and $y$ is an invariant coordinate with respect to the corresponding group action, we would have $\expval{\grad y,X}=0$. We can enforce this condition on the tangent space of $\mcM$ (as a soft constraint) using \cref{alg:CAE_proj}, to computationally infer invariant coordinates, as demonstrated in the simple example that follows. This process can be seen as a form of \textit{canonicalization} \cite{olver2024equivariant,aslan2023group_eq,kaba2023equivariance}, where, we find a submanifold that intersects each orbit with respect to the group \textit{once} (see e.g. \cref{fig:ex_3_S_curve_k_means}(a)). This submanifold is parametrized by the invariants of the group action, and can be used to facilitate the computation of invariant and equivariant functions. It is not, however, guaranteed to exist globally under general group actions.

\label{sec:symmetries_exmp}
\begin{exmp}[S-curve]
\label{exmp:s_curve_symm}
    \begin{figure*}[ht]
        \centering
	\begin{subfigure}[b]{0.30\textwidth}
		\includegraphics[width=\textwidth]{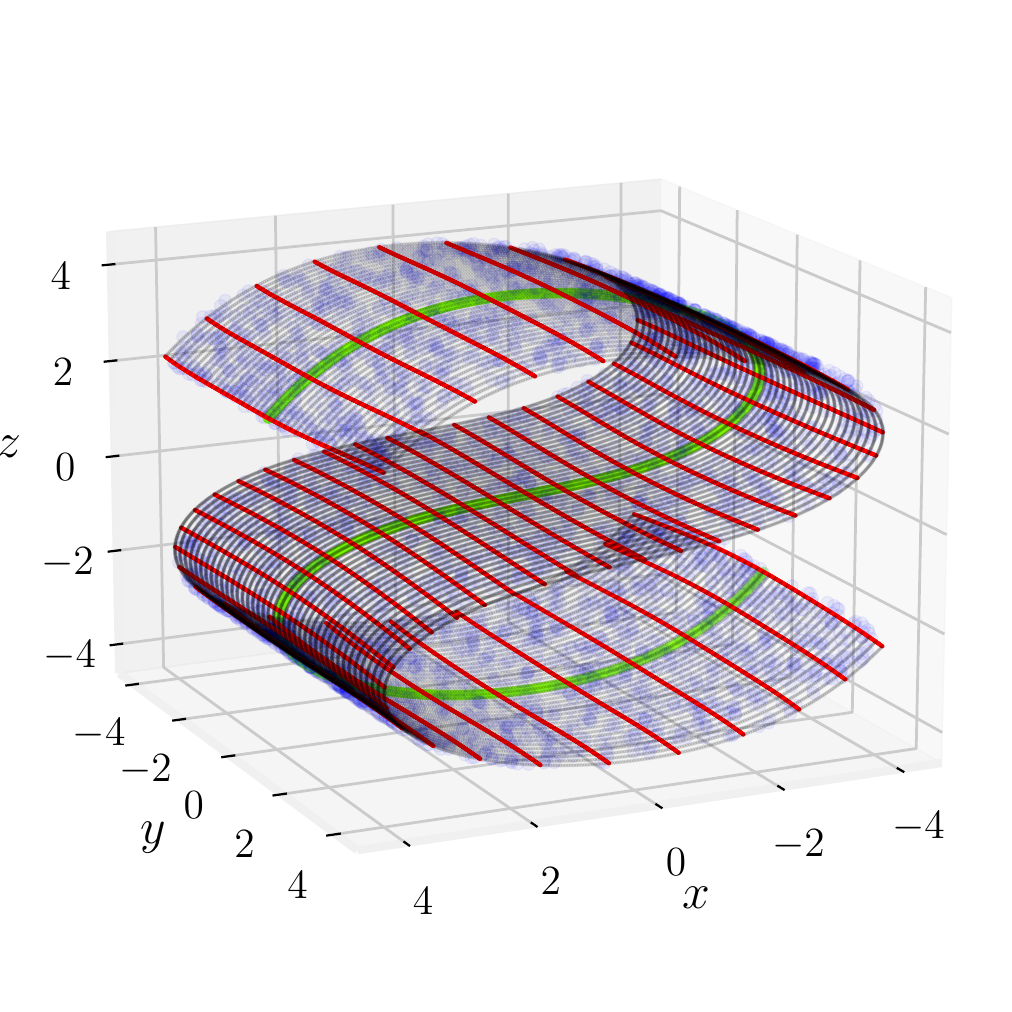}
		\caption{}
		\label{fig:s_curve_k_means_level_sets}
	\end{subfigure}
	\begin{subfigure}[b]{0.30\textwidth}\includegraphics[width=\textwidth]{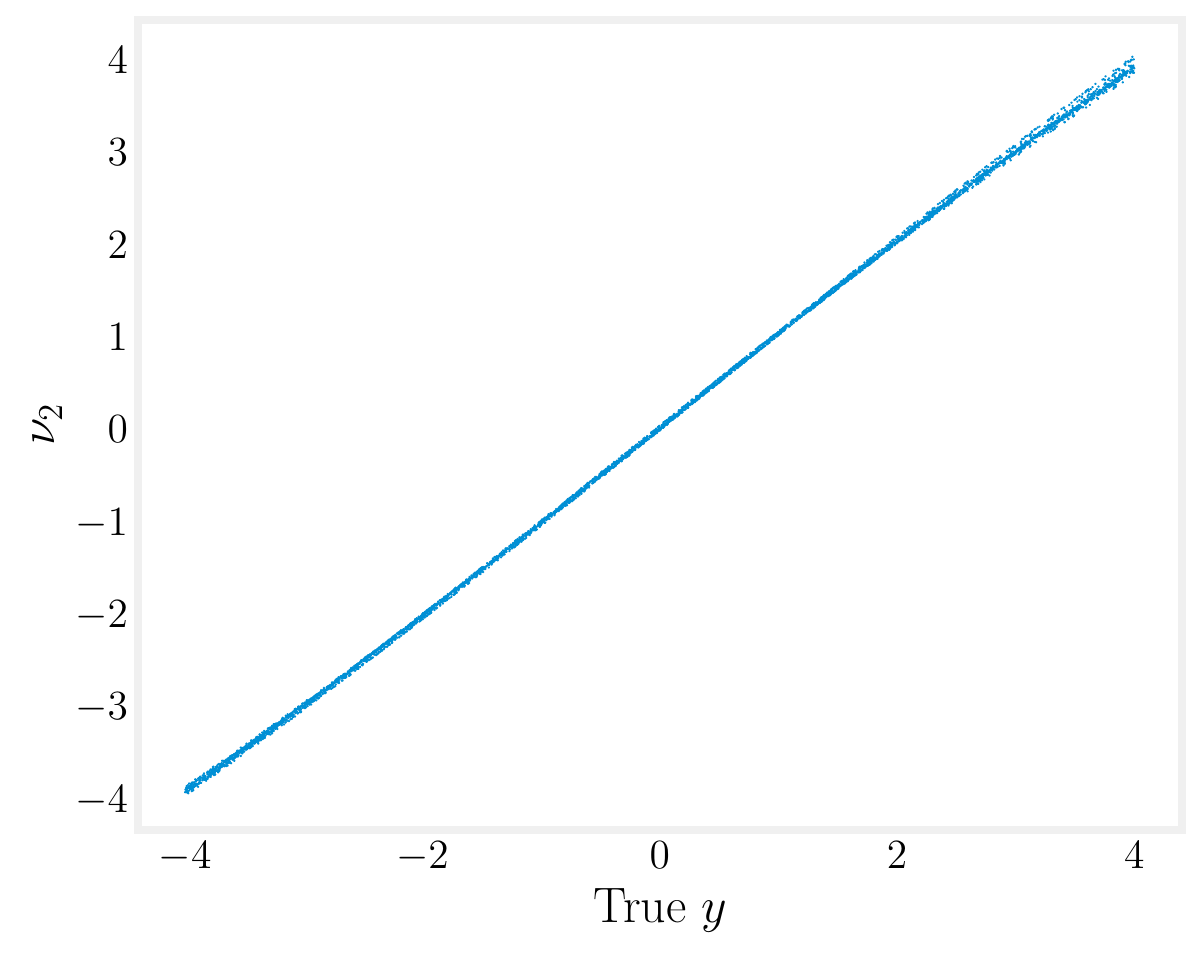}
		\caption{}
		\label{fig:s_curve_k_means_embedding_nu_1}
	\end{subfigure}
	\begin{subfigure}[b]{0.30\textwidth}\includegraphics[width=\textwidth]{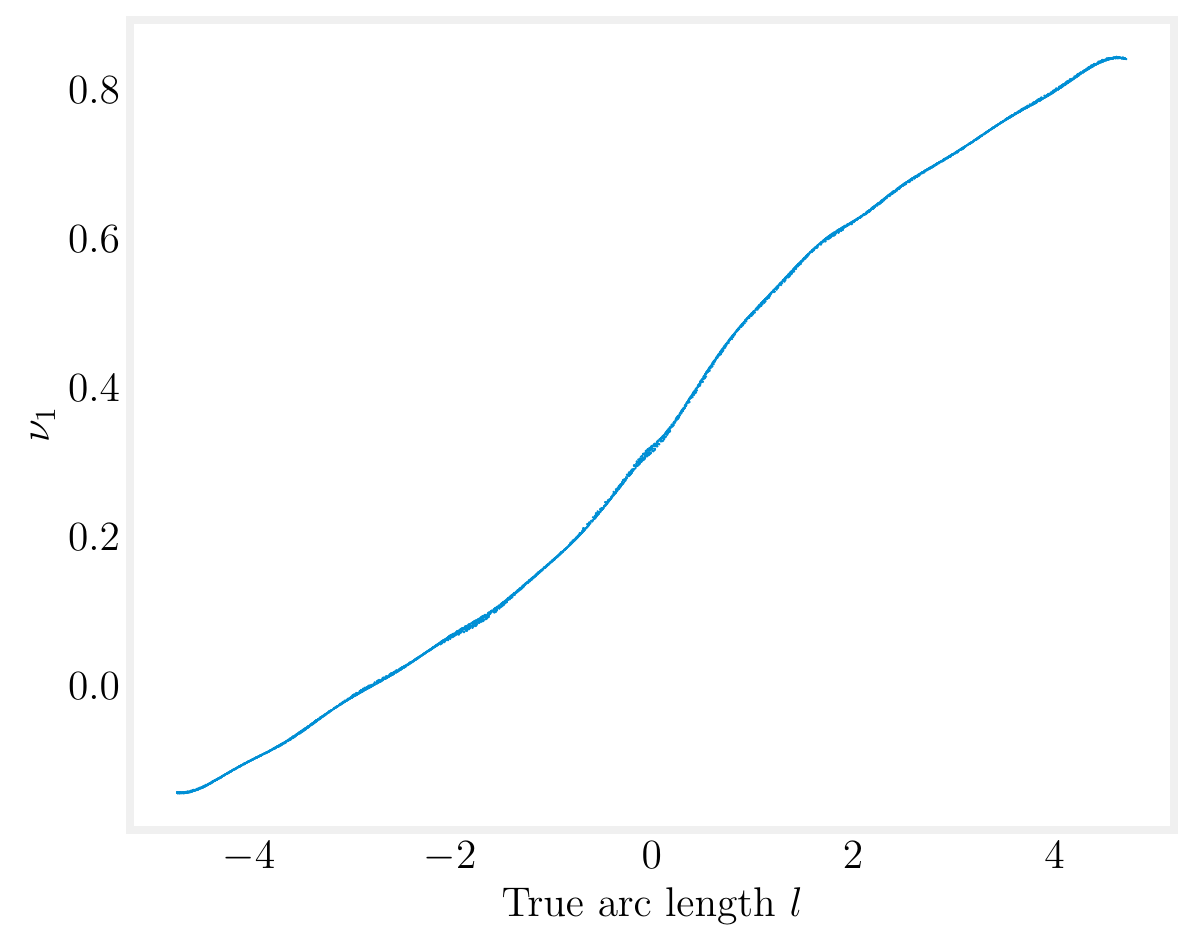}
		\caption{}
		\label{fig:s_curve_k_means_embedding_nu_2}
	\end{subfigure}
	\caption{Training result of \cref{alg:CAE_proj} on the S-curve data set, when $y$ is prescribed as a latent variable. In \cref{fig:s_curve_k_means_level_sets} we plot the level sets of the prescribed $y$-coordinate (\textcolor{red}{red}), and the level sets of the inferred ``arc length" coordinate (black, \textcolor{green} {green}). \cref{fig:s_curve_k_means_embedding_nu_1,fig:s_curve_k_means_embedding_nu_2} show that the true and network inferred coordinates are one-to-one.}
	\label{fig:ex_3_S_curve_k_means}
\end{figure*}

 We briefly demonstrate an application of Algorithm 2, where  invariance to a symmetry group on the submanifold sampled by the data set is imposed by locally projecting on the tangent space $T\mcM$.
 For the two-dimensional S-curve data set, we consider the projection on the $y$-coordinate of each point as a known latent variable. 
 One may locally think of the one-parameter Lie group whose action is generated by the associated vector field $\pdv{y}$ on $T\mcM$. This corresponds to translations in the direction of the $y$-axis.
 We desire an orthogonal parametrization of the manifold in which the second latent variable satisfies $\nu_2\approx y$, while the first ($\nu_1$) remains invariant along $y$. In this case, we treat the intrinsic dimension $k=2$ as known. In order to enforce invariance on $T\mcM$ we assign each point to a local cluster of neighbors whose principal components we compute. At each such point, we project the (NN-generated) latent-variable gradients on the plane spanned by these local principal components, and subsequently compute the orthogonality loss. In \cref{fig:ex_3_S_curve_k_means} we show the result of a single optimization run of \cref{alg:CAE_proj}. Following the notation from \cref{sec:Numerical_Method} we define the loss function
\begin{equation}
    \mcL = \frac{1}{N}\sum_{i=1}^N\norm{\vb{x}_i-\hat{\vb{x}}_i}_2^2+\frac{1}{N}\sum_{i=1}^N\abs{\expval{\grad^\mcM\nu_1({\vb{x}_i}),\grad^\mcM\nu_2({\vb{x}_i})}}+\frac{1}{N}\sum_{i=1}^N\norm{\vb{y}-\vb{\nu}_{2_i}}_2^2.
\end{equation}
We note that, while here we depict a successful optimization result, the algorithm may produce a ``patchy" chart, similar to that of \cref{fig:ex_2_training} in the Appendix, where the chart `jumps' connecting non-neighboring segments of the surface. This is due to the large extrinsic curvature of the embedded data, which do not combine well with the generic network initialization (normally distributed random weights) we use.

\end{exmp}

\section{Summary and Discussion}
\label{sec:Discussion}
Throughout this work, we develop a framework in which a single autoencoder can simultaneously infer the dimension of,  and produce a minimal chart for, a sampled nonlinear submanifold $\mcM$ embedded in Euclidean space. 
In \cref{sec:Theory_and_Methodology} we outline the theoretical context and implementation based on enforcing orthogonality between the discovered latent components (\cref{eqn:CAE_loss}). 
Given the readily available estimates of neural network gradients in ambient space due to Automatic Differentiation, we implement \cref{alg:CAE} on several data sets in \cref{sec:Numerical_Examples}. While the orthogonality constraint in this case does not \textit{guarantee} a `correct' estimate of the dimension, the implicit bias given to the network, coupled with the optimization dynamics, produces useful and fast results on our computational examples. Here, we empirically observe in \cref{fig:ex_1_grads}, and the corresponding figures of all other examples, that it is common for the latent dimension to increase \textit{sequentially during training}. 

Despite the lack of guarantees, we note that this method may be useful in cases where a high ambient dimension makes other nonlinear methods (e.g. Diffusion Maps) harder to use, and a `correct' intrinsic dimensionality answer is not generically reachable by simple optimization means. 

We further develop the capability to work directly on the tangent space $T\mcM$ of a given data set in \cref{alg:CAE_orthog,alg:CAE_proj}, a combination of which is implemented in \cref{sec:symmetries_exmp}. Because at each point $p\in\mcM$, the tangent space is estimated using a local, linear dimension reduction technique (such as PCA), the dimension of the latent space need not be inferred. However, due to the relationship of orthogonality and invariance, our ability to work on $T\mcM$ can be used to leverage the approximation power of neural networks to produce data-driven invariant functions (defined either in ambient space or on $\mcM$). 

The main advantage of our method, finally, is that it may circumvent a two-step approach, applying first a graph-based algorithm to infer latent dimensionality and coordinates for a fixed set of points,  followed by training an autoencoder to estimate the associated continuous embedding maps. Additionally, the relation of orthogonality to invariance (in the case of smooth group actions on smooth manifolds) may be useful, due to the simplicity with which orthogonality constraints on $T\mcM$ (or ambient space) can describe more complicated relationships (i.e. differential equations) on nonlinear manifolds. 

\bibliography{main.bib}
\bibliographystyle{abbrv}


\clearpage

\appendix

\section{Algorithms}
\label{app:Algs}

We give a short description of the computational steps involved in the CAE optimization described in \cref{sec:Theory_and_Methodology}. The only nontrivial step is that involving the inner product computation, which is possible through automatic differentiation. In practice, we use more sophisticated algorithms than simple gradient descent to perform the backward pass (e.g. Adam).

\begin{algorithm}
	\caption{CAE Dimension Reduction}
	\label{alg:CAE}
	\KwData{ambient dimension ($n\in\N$), $k$-dimensional data samples $\qty{\vb{x}_i}_{i=1}^N$ embedded in $\R^n$}
	\KwResult{estimated latent dimension $k\in\N$, chart $\psi$ and inverse over data set}
	Set the latent layer dimension equal to $n$. Randomly initialize encoder $\mathfrak{e}$ and decoder $\mathfrak{d}$ weights ($w_\mathfrak{e},w_\mathfrak{d}$ respectively). Set the learning rate $\eta$ and error tolerance $\epsilon$ to be small positive constants. Set $\alpha\in\R$ to be a positive constant\\
	\While{$\mathcal{E}_\text{CAE}\geq \epsilon$}{
		// reconstruction forward pass:
		\begin{gather}
			\qty{\bm{\nu}_i}_{i=1}^N=\mathfrak{e}(\qty{\vb{x}_i}_{i=1}^N)\\
			\qty{\hat{\vb{x}}_i}_{i=1}^N=\mathfrak{d}(\qty{\bm{\nu}_i}_{i=1}^N)
		\end{gather}
		// compute reconstruction loss:
		\begin{equation}
			\label{eqn_alg:recon}
			\mathcal{E}_\text{CAE}=\frac{1}{N}\sum_{i=1}^N\norm{\vb{x}_i-\hat{\vb{x}}_i}_2^2
		\end{equation}
		// compute orthogonality loss, with $\grad\nu_j({\vb{x}})$ computed with automatic differentiation
		\begin{equation}
			\label{eqn_alg:ortho}
			\mathcal{E}_\text{CAE}\mathrel{+}=\alpha\frac{1}{N}\sum_{i=1}^N\sum_{j> k}\abs{\expval{\grad\nu_j({\vb{x}}),\grad\nu_k({\vb{x}})}}^2
		\end{equation}
		// perform backward pass:
		\begin{align}
			w_\mathfrak{e}\mathrel{-}=\eta\grad_{w_\mathfrak{e}}\mathcal{E}_\text{CAE}\\
			w_\mathfrak{d}\mathrel{-}=\eta\grad_{w_\mathfrak{d}}\mathcal{E}_\text{CAE}
		\end{align}
	}
\end{algorithm}

It is clear that one may replace the $\ell^2$ reconstruction norm (\cref{eqn_alg:recon}) and the $\ell^2$ inner product (\cref{eqn_alg:ortho}) with any other suitable candidates, which may or may not be application specific. Instead of an inner product as a measure of orthogonality, one would ideally like to use the cosine of the angle between vectors (i.e. the normalized inner product). The latter is a stable measure of orthogonality, and works successfully when the correct dimension of the latent space is known (as in \cite{parameters}). However, it is not defined at the origin, and does not allow an iterative algorithm to make components smaller (eventually tending to zero), thus interpolating between maps of different dimension. That makes inner products suitable for the current application. One may additionally increase the value of $\alpha$ or use an $\ell^1$ norm for orthogonality to encourage lower-dimensional latent spaces.

A second algorithm can make use of automatic differentiation and subsequent projection onto the tangent space of the data manifold $\mcM$. The tangent space at each point can be estimated by performing PCA on the $k$-nearest neighbors at each point $p\in\mcM\subset\R^n$ and this can be done as a preprocessing step. This adds a single projection step to our previous optimization procedure, summarized in \cref{alg:CAE_proj}.

\begin{algorithm}
	\caption{CAE Dimension Reduction with gradient projection}
	\label{alg:CAE_proj}
	\KwData{ambient dimension ($n\in\N$), $k$-dimensional data sample $\qty{\vb{x}_i}_{i=1}^N$ embedded in $\R^n$}
	\KwResult{true latent dimension $k\in\N$, orthogonal chart $\psi$ and inverse over data set}
	Set the latent layer dimension equal to $n$. Randomly initialize encoder $\mathfrak{e}$ and decoder $\mathfrak{d}$ weights ($w_\mathfrak{e},w_\mathfrak{d}$ respectively). Set the learning rate $\eta$ and error tolerance $\epsilon$ to be small positive constants. Set $\alpha\in\R$ to be a positive constant. Set $k_\text{NN}$ to be the number of nearest neighbors considered for each point.\\
	\For{each $\vb{x}_i$}{
		Compute the $k_\text{NN}$ points closest to $\vb{x}_i$.\\
		Perform PCA and parametrize the tangent space $T_{\vb{x}_i}\mcM$ using its leading principal components.
	}
	\While{$\mathcal{E}_\text{CAE}\geq \epsilon$}{
		reconstruction forward pass:
		\begin{gather}
			\qty{\bm{\nu}_i}_{i=1}^N=\mathfrak{e}(\qty{\vb{x}_i}_{i=1}^N)\\
			\qty{\hat{\vb{x}}_i}_{i=1}^N=\mathfrak{d}(\qty{\bm{\nu}_i}_{i=1}^N)
		\end{gather}
		compute reconstruction loss:
		\begin{equation}
			\mathcal{E}_\text{CAE}=\frac{1}{N}\sum_{i=1}^N\norm{\vb{x}_i-\hat{\vb{x}}_i}_2^2
		\end{equation}
		compute gradients and project onto $T_{\vb{x}_i}\mcM$
		\begin{equation}
			\grad^\mcM\nu_i({\vb{x}})=\text{proj}_{T_{\vb{x}\mcM}}\grad\nu_i({\vb{x}})\quad\forall i
		\end{equation}
		compute orthogonality loss:
		\begin{equation}
			\mathcal{E}_\text{CAE}+=\alpha\frac{1}{N}\sum_{i=1}^N\sum_{j>k}\abs{\expval{\grad^\mcM\nu_j({\vb{x}}),\grad^\mcM\nu_k({\vb{x}})}}^2
		\end{equation}
		perform backward pass:
		\begin{align}
			w_\mathfrak{e}-=\eta\grad_{w_\mathfrak{e}}\mathcal{E}_\text{CAE}\\
			w_\mathfrak{d}-=\eta\grad_{w_\mathfrak{d}}\mathcal{E}_\text{CAE}
		\end{align}
	}
\end{algorithm}

We note that there may be multiple admissible ways of estimating the local tangent space and thus estimate of $\grad^\mcM$ (e.g. one way alternatively define a scale parameter $\tau$ and define neighbors of a point $p\in\mcM$ as the set $\qty{p'\in\mcM:\norm{p-p'}\leq\tau}$). However, in making the decision to project, the latent layer dimension is fixed by the process followed, and so defining an autoencoder with additional latent components becomes unnecessary. This process would still provide an embedding map (and its inverse) for the manifold at hand.

Finally, we observe that \cref{alg:CAE_proj} produces a chart that is orthogonal \textit{on} $T_\mcM$, while \cref{alg:CAE} may not, since orthogonality is generally not preserved when projecting. To address that issue, we are capable of \textit{a posteriori} orthogonalizing a chart on $T\mcM$, by computing the tangent vectors in the ambient space using automatic differentiation of the decoder network $\mathfrak{d}$. This can only be done \textit{after} a chart of the correct dimension is learned, and its use is more so in cases where a particular latent parameter may be meaningful and `disentangling it' from other latent parameters may be useful in terms of interpretability (as in \cite{parameters}). This process is summarized in \cref{alg:CAE_orthog}.

\begin{algorithm}
	\caption{CAE a posteriori orthogonalization}
	\label{alg:CAE_orthog}
	\KwData{$k$-dimensional data sample $\qty{\vb{x}_i}_{i=1}^N$ embedded in $\R^n$, encoder and decoder networks $(\mathfrak{e},\mathfrak{d})$}
	Set the learning rate $\eta$ and error tolerance $\epsilon$ to be small positive constants. Set $\alpha\in\R$ to be a positive constant.
	\KwResult{Conformal embedding of $\mcM$ of $k$-dimensional data manifold in $\R^n$}
	\While{$\mathcal{E}_\text{CAE}\geq \epsilon$}{
		reconstruction forward pass:
		\begin{gather}
			\qty{\nu_i}_{i=1}^N=\mathfrak{e}(\qty{\bm{x}_i}_{i=1}^N)\\
			\qty{\hat{\vb{x}}_i}_{i=1}^N=\mathfrak{d}(\qty{\bm{\nu}_i}_{i=1}^N)
		\end{gather}
		compute reconstruction loss:
		\begin{equation}
			\mathcal{E}_\text{CAE}=\frac{1}{N}\sum_{i=1}^N\norm{\vb{x}_i-\hat{\vb{x}}_i}_2^2
		\end{equation}
		compute orthogonality loss:
		\begin{equation}
			\mathcal{E}_\text{CAE}+=\alpha\frac{1}{N}\sum_{i=1}^N\sum_{j> k}\abs{\expval{D\hat{\vb{x}}_j({\bm{\nu}_i}),D\hat{\vb{x}}_k({\bm{\nu}_i})}}^2
		\end{equation}
		perform backward pass:
		\begin{align}
			w_\mathfrak{e}-=\eta\grad_{w_\mathfrak{e}}\mathcal{E}_\text{CAE}\\
			w_\mathfrak{d}-=\eta\grad_{w_\mathfrak{d}}\mathcal{E}_\text{CAE}
		\end{align}
	}
\end{algorithm}
We denote by $D$ the gradient of the outputs of the decoder $\mfd$ with respect to the latent variables $\bm{\nu}$

\clearpage

\section{Network Architectures}
\label{app:architectures}

\begin{table*}[h!]
    \centering
    \begin{tabular}{lccccc}
        {} &Depth & Width & \multicolumn{3}{c}{Activation}\\
        \midrule
        \cref{exmp:toy} (Toy) & 5  & 10 & \multicolumn{3}{c}{$\tanh$ (all layers)}\\
        \cref{exmp:circle} (Circle) & 7 & 10 & \multicolumn{3}{c}{$\tanh$ (1-5), none (6,7)}\\
        \cref{exmp:S_curve} (S-curve) & 7 & 10 &\multicolumn{3}{c}{$\tanh$ (1,3,5), $\text{hardtanh}$ (2,4), none (6-7)}\\
        \cref{exmp:KS} (KS) & 5 & 20 & \multicolumn{3}{c}{$\tanh$ (all layers)}\\
        \cref{exmp:CI} (CI) & 5 & 20 & \multicolumn{3}{c}{$\tanh$ (all layers)}\\
    \end{tabular}
    \caption{Architecture specifications for the networks used in each example of section \cref{sec:Numerical_Examples}}
    \label{tab:architectures}
\end{table*}

The basic building block of an autoencoder is a feed-forward neural network:

\begin{definition}[Feed-Forward Network]\label{definition:FFNN}
	A single \textit{layer} of a feed forward network is a function of the input $\vb{x}\in\R^n$ of the form $f\vb{x}=\rho(W\vb{x}+b)$ where $W\in\R^m\times\R^n$ is a linear transformation (and $m$ is the width of the layer), $b\in\R^m$ is a bias term, and $\rho$ is a nonlinear `activation function' applied point-wise. The image of each layer lies in $\R^m$. A \textit{feed-forward network} of $L$ feed-forward layers is a function that applies single layers to its input recursively. Letting \(\qty{W_i,b_i,\rho_i}_{i=1}^L=\qty{f_i}_{i=1}^L\) specify each $i$-th layer, it can be expressed as $f\vb{x}=f_L\circ f_{L-1}\circ...\circ f_1\vb{x}$. Note that layer widths must be consistent.
\end{definition}

The collection of weights and biases are often denoted by $\theta=\qty{W_i,b_i}_{i=1}^L$, specifying the corresponding network $f_\theta$. In our work, these parameters are initialized at random (using the standard \texttt{pytorch} initialization) and are optimized using Adam. These are generic choices that may be adapted to better suit particular applications.

It is important to note that feed-forward networks where $\rho\in C^\infty$ are themselves $C^\infty$ functions, yielding (asymptotically) a family of universal smooth function approximators (\cref{sec:concerning_approx}).

The specifications for the architecutes used in each example of \cref{sec:Numerical_Examples} are listed in \cref{tab:architectures}.

\section{Additional Numerical Examples}
\label{app:numerical_examples}

\begin{table*}[h!]
    \centering
    \begin{tabular}{lcccccc}
        \multirow{2}{*}[-2em]{} & \multicolumn{3}{c}{Error} & \multicolumn{3}{c}{Dimension}\\
        \addlinespace[2pt]
        \cmidrule(lr){2-4} \cmidrule(lr){5-7} \\
        {} &Training $(\mcL_\text{CAE})$ & Test $(L^2)$ & Scale & Ambient  & Intrinsic & Inferred\\
        \midrule
        \cref{exmp:toy} (Toy) & \num{1.6e-4}  & \num{1.5e-4} & $[0,1]^3$ & 3  & 2 & 2 \\
        \cref{exmp:circle} (Circle) & \num{2.0e-4} & n/a & $[-1, 1]^2$ & 2 & $1^*$ & 1 \\
        \cref{exmp:S_curve} (S-curve) & \num{1.7e-2}& \num{1.6e-2} & $[-4,4]^3$ & 3 & 2 & 2 \\
        \cref{exmp:KS} (KS) & \num{2.9e-4} & \num{3.0e-4}& $[0,1]^8$ & 8 & 3 & 3 \\
        \cref{exmp:CI} (CI) & \num{6.7e-4} &\num{7.0e-4}& $[0,1]^{10}$ & 10 & 2 & 2 \\
    \end{tabular}
    \caption{Summary of the dimension reduction results for the examples of \cref{sec:Numerical_Examples}. The test error is computed after training, where the `unused' latent features are set to their mean during training. $^*$ The tangent space for the circle is locally one dimensional, even though it cannot be embedded $\R$.}
    \label{tab:numerical_errors}
\end{table*}

\subsection{Dimension Reduction on Toy Data Sets}

\begin{exmp}[Circle]\label{exmp:circle}

        \begin{figure*}
		\centering
		\includegraphics[width=0.65\textwidth]{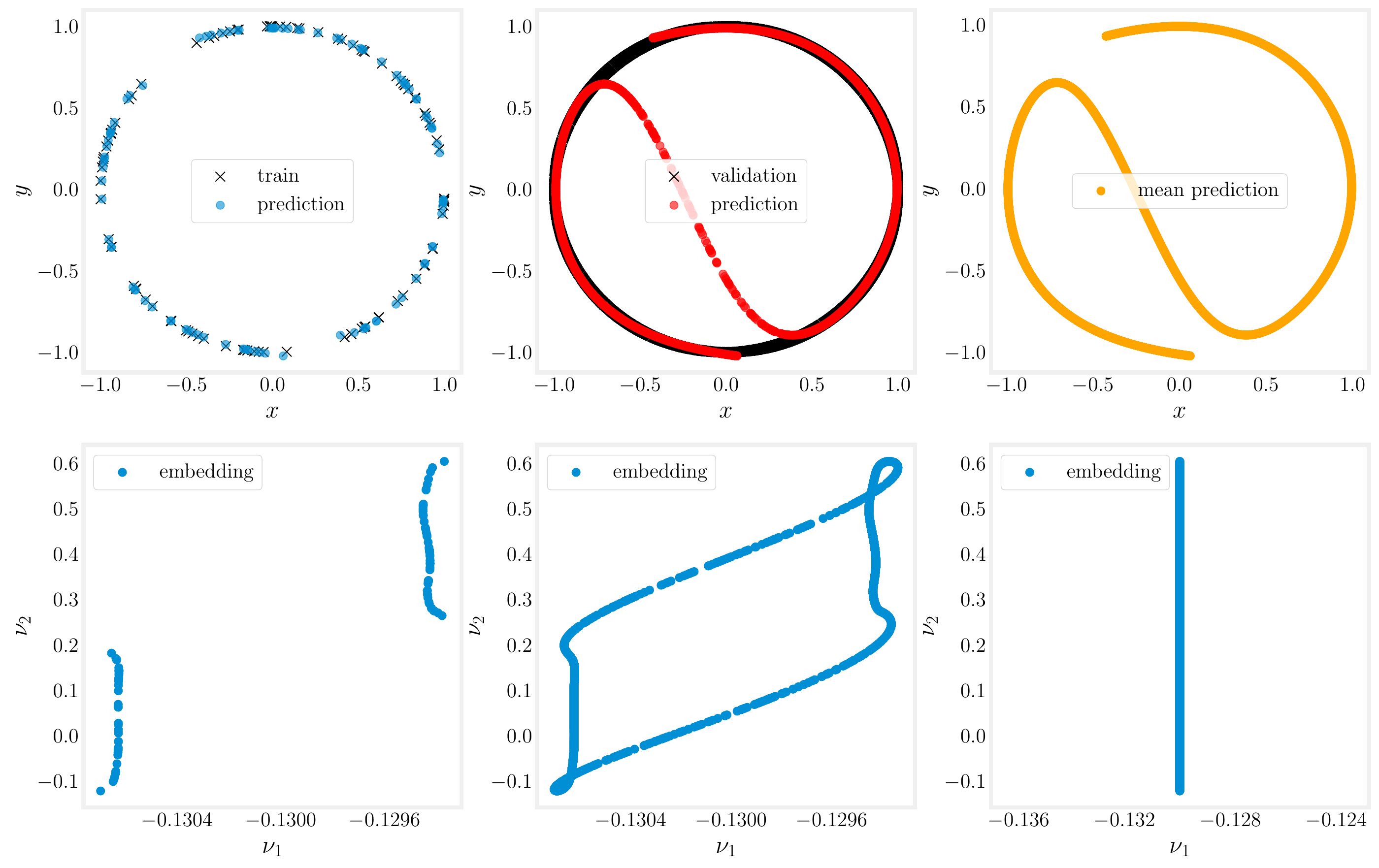}
		\caption{Visualization of the validation step in \cref{exmp:circle}. The top row depicts data in ambient space ($\R^2$) while the bottom row depicts the corresponding embedding in latent space. For the first two columns, the latent-space embedding is produced by the encoder on the train or validation set. For the third column, the latent space is constructed manually and the prediction is made using the trained decoder.}
		\label{fig:ex_2_validation}
	\end{figure*}

    \begin{figure*}
		\begin{subfigure}[b]{0.45\textwidth}
			\includegraphics[width=\textwidth]{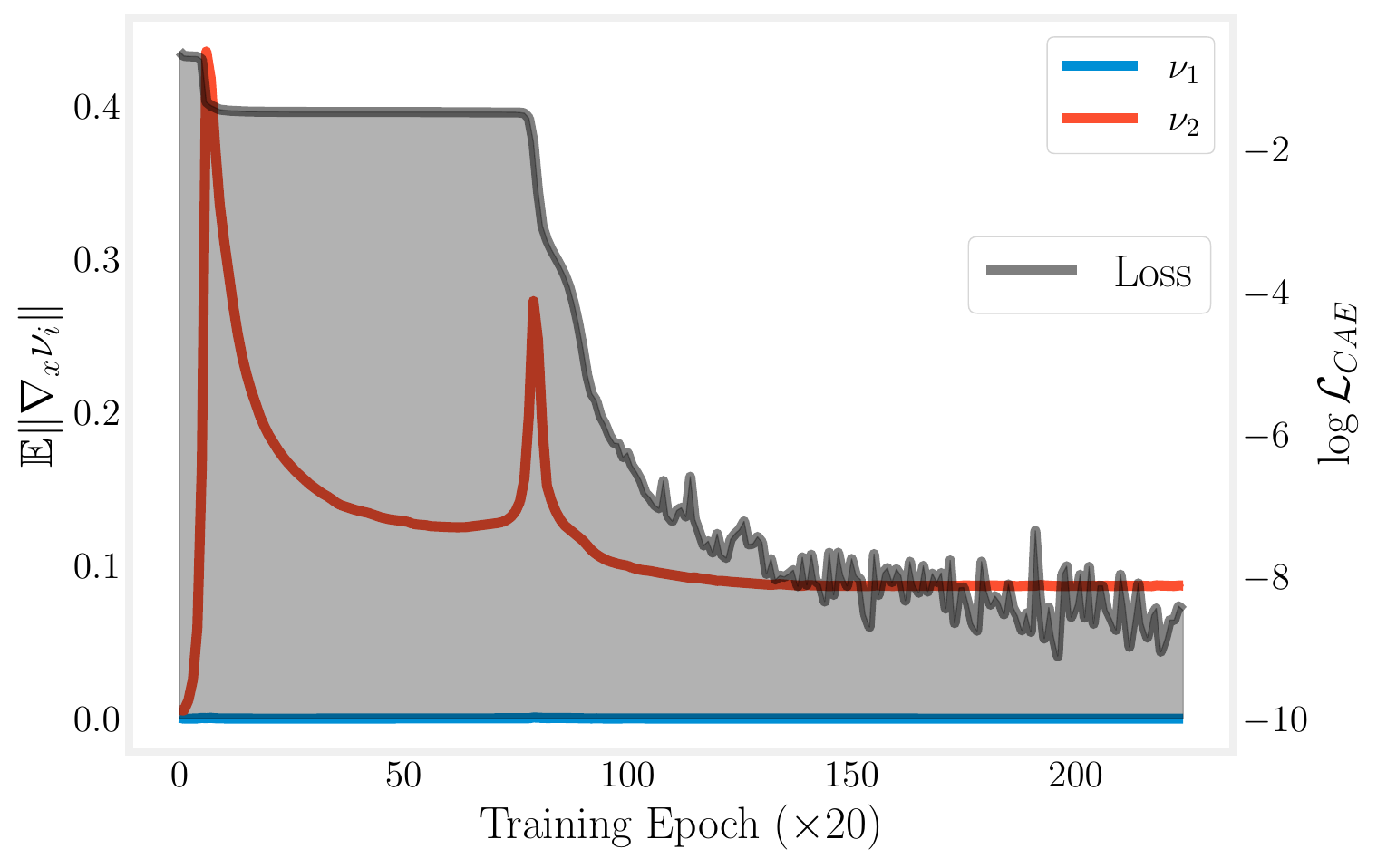}
			\caption{}
			\label{fig:ex_2_circle_grads}
		\end{subfigure}
		\begin{subfigure}[b]{0.45\textwidth}\includegraphics[width=\textwidth]{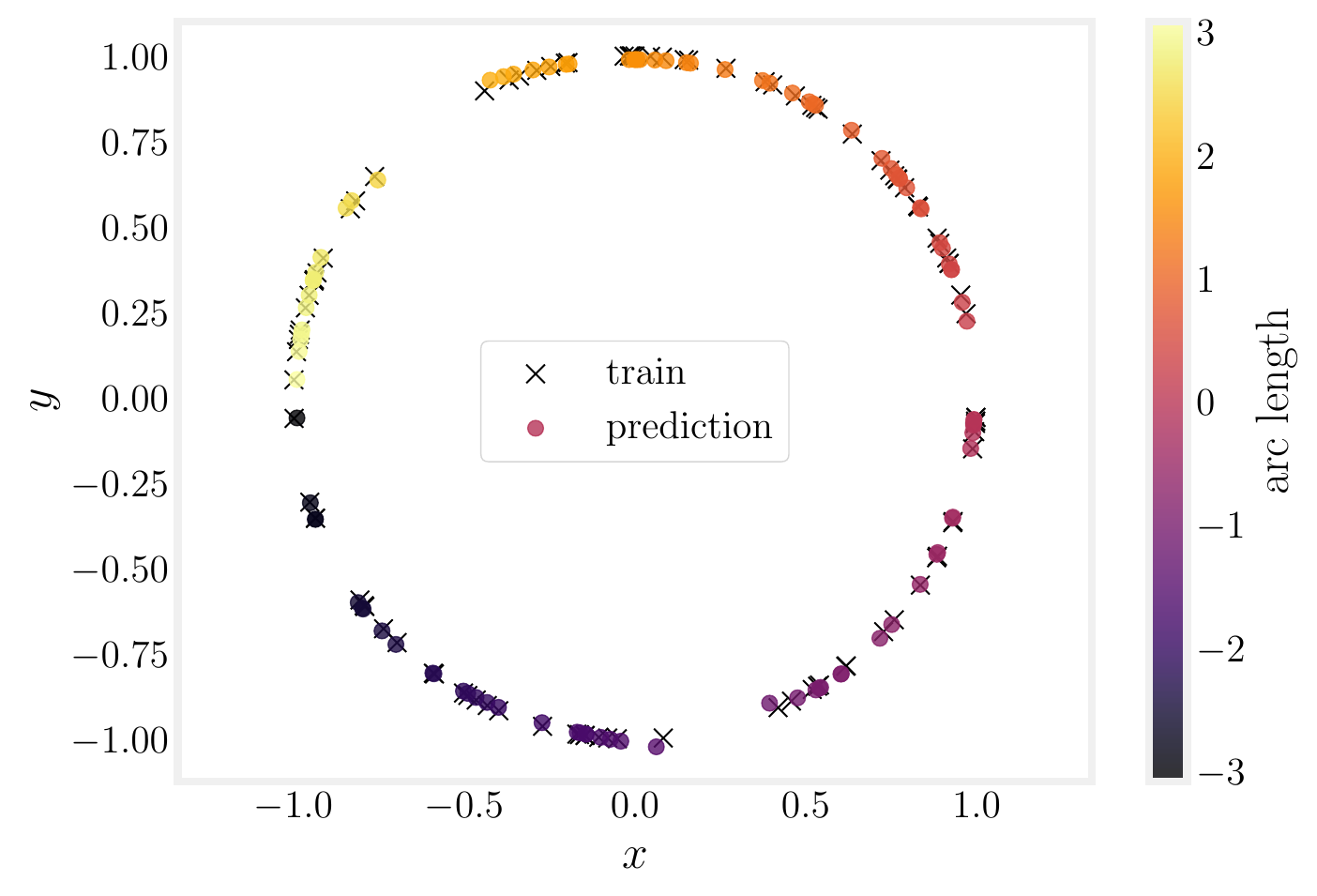}
			\caption{}
			\label{fig:ex_2_vals}
		\end{subfigure}
		\caption{Optimization result for a single run of \cref{alg:CAE} on the circle data set of \cref{exmp:circle}. \cref{fig:ex_2_circle_grads} (left) is similar to \cref{fig:ex_1_grads} of \cref{exmp:toy}. \cref{fig:ex_2_vals} (right) shows both the training set and its reconstruction by the autoencoder colored by $\arctan_2(y/x)$ in ambient space ($\R^2$).}
		\label{fig:ex_2_training}
    \end{figure*}

    It is instructive to look at the behavior of the proposed algorithm when there are topological obstructions to the single-chart assumption. The circle (as embedded in $\R^2$) is a one dimensional manifold that cannot be covered by a single chart. Furthermore, any smooth function that parametrizes it’s arclength must have a singularity somewhere inside the circle, it must be diffeomorphic to
    \begin{equation}
		\theta=\arctan_2\qty(\frac{y}{x})
    \end{equation}
    which is impossible for a $C^\infty$ network to express. We proceed as follows:
	
	We sample $N=100$ points uniformly at random from $S^1\subset\R^2$ as a training set and apply \cref{alg:CAE} with an $L_1$ loss term in the orthogonality component (\cref{eqn:CAE_loss}). In \cref{fig:ex_2_training} we see that the architecture `correctly' identifies that the data set parametrization can be one-dimensional, with a training error of $\mcL_\text{CAE}=\num{2e-4}$. However, once we obtain a dense sample of the circle as a validation step as in \cref{fig:ex_2_validation}, we observe that the autoencoder {\em fails} to properly reconstruct the circle despite its success on the training set. In the first column we observe that the embedding of the training set is indeed one-dimensional (note that the scale of the $x$-axis is very small compared to $y$-, but the jump between the components is already indicative of an irregularity). In the second column, the  embedding of a dense circle (black) is visibly a closed curve in latent space, but the autoencoder fails to reconstruct it and instead produces the `irregular' $S$-curve (\textcolor{red}{red}). Finally, in the third column we generate a one-dimensional (with $\nu_1=\E[\nu_1^\text{train}]$) latent space, and confirm that its image is another irregular $S$-shaped curve that interpolates the circle well (only) in a neighborhood of the training data.
	
	Of course, since the circle is not embeddable in one dimension, it is to be expected that the network should fail. It is still important that the algorithm is able to identify the dimension of the tangent space locally, which may be useful in downstream optimization tasks in applications. We further note that the irregularity of the inferred $S$-shaped-curve corresponds to large (extrinsic) curvature in ambient space and a large Lipschitz constant of the decoder (small distances in latent space become large in ambient space). Firstly, such irregularity can be reduced, conceptually, by establishing control of the Lipschitz constants of the encoding and decoding networks, producing more regular embeddings. While we do not control the Lipschitz constants in our architectures, it is possible to do so, and implementations of such constraints is an active area of research \cite{fazlyab2019lip,pauli2021lip}. Alternatively, replacing the deterministic CAE architecture with a VAE could also enforce convexity of the latent embedding, due to its ability to generate perturbed points in latent space. Secondly, studying the presence of such irregularities can be used to \textit{infer} global topological issues that may not be known {\em a priori}, giving us crucial information about particular data sets when such characteristics are important.
	
\end{exmp}

\begin{exmp}[S-Curve - Revisited]\label{exmp:S_curve}

\begin{figure*}[h!]
    \centering
    \includegraphics[width=0.85\textwidth]{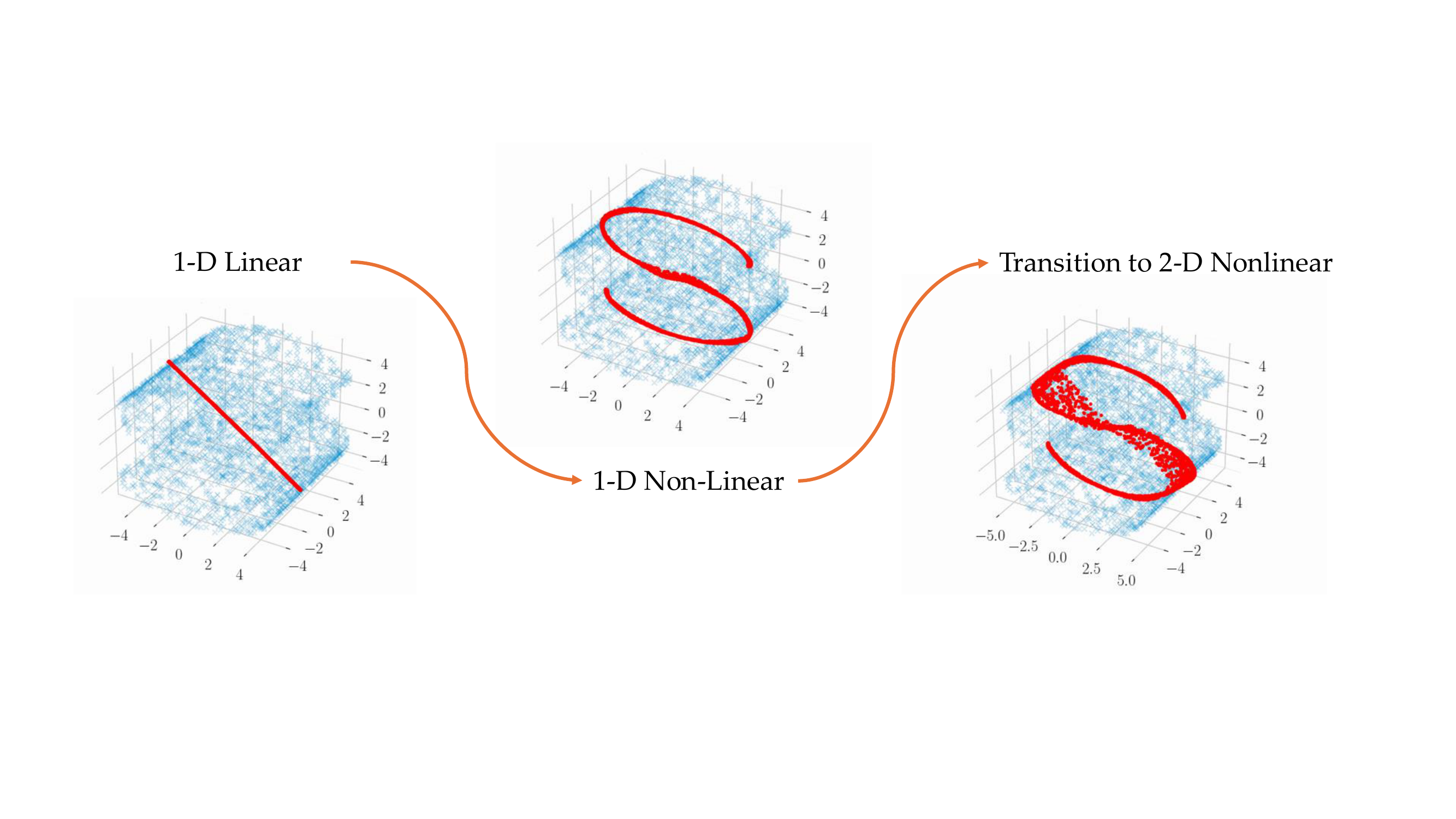}
    \caption{Embeddings produced for the S-curve data set during training using \cref{alg:CAE}. The embeddings increase in complexity sequentially, where we whitness a transition between a 1-dimensional linear approximation which becomes nonlinear, before becoming two-dimensional}
    \label{fig:learning_transitions}
\end{figure*}

\begin{figure*}[h!]
		\begin{subfigure}[b]{0.33\textwidth}
			\includegraphics[width=\textwidth]{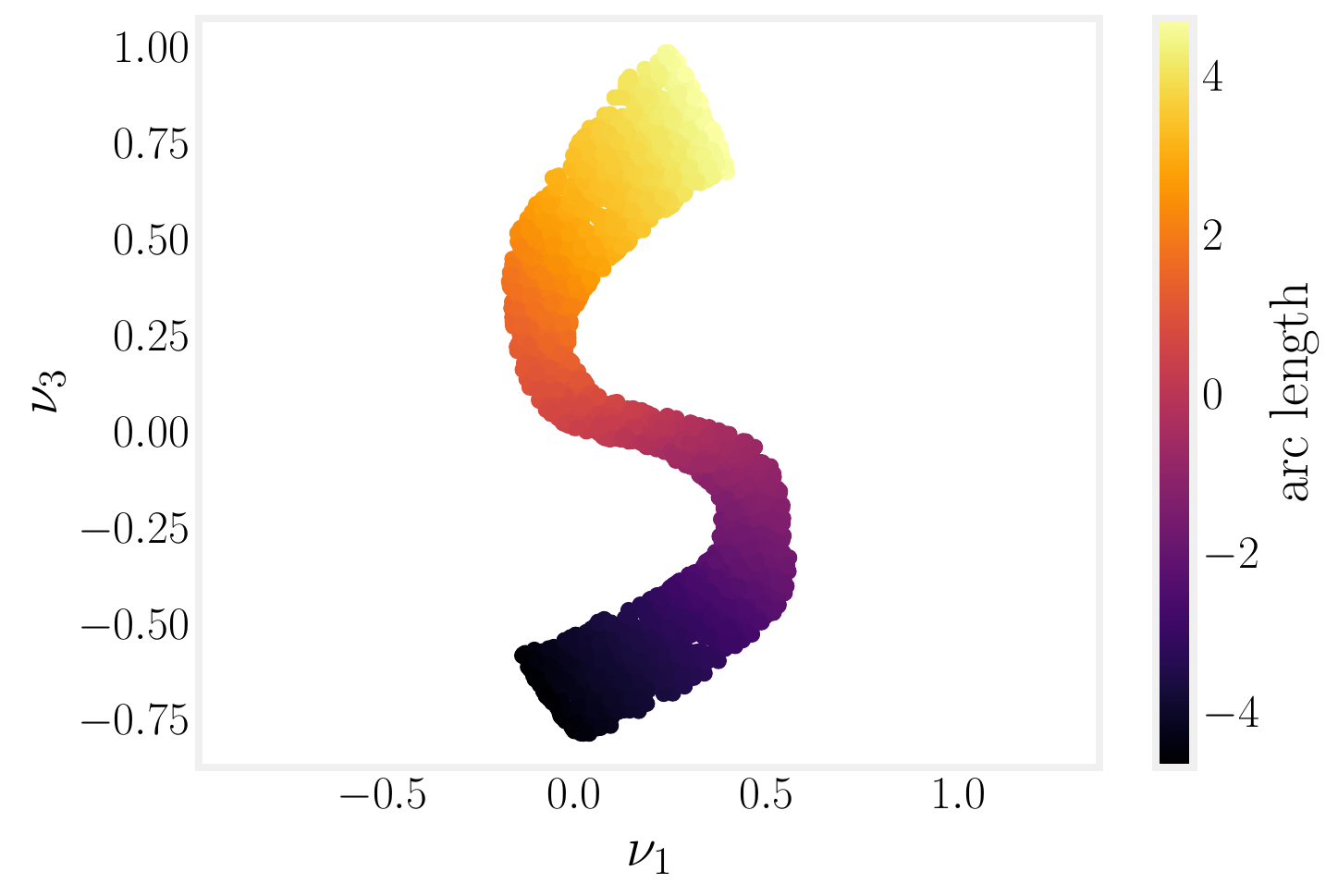}
			\caption{}
			\label{fig:ex_3_s_curve_latent}
		\end{subfigure}
		\begin{subfigure}[b]{0.33\textwidth}\includegraphics[width=\textwidth]{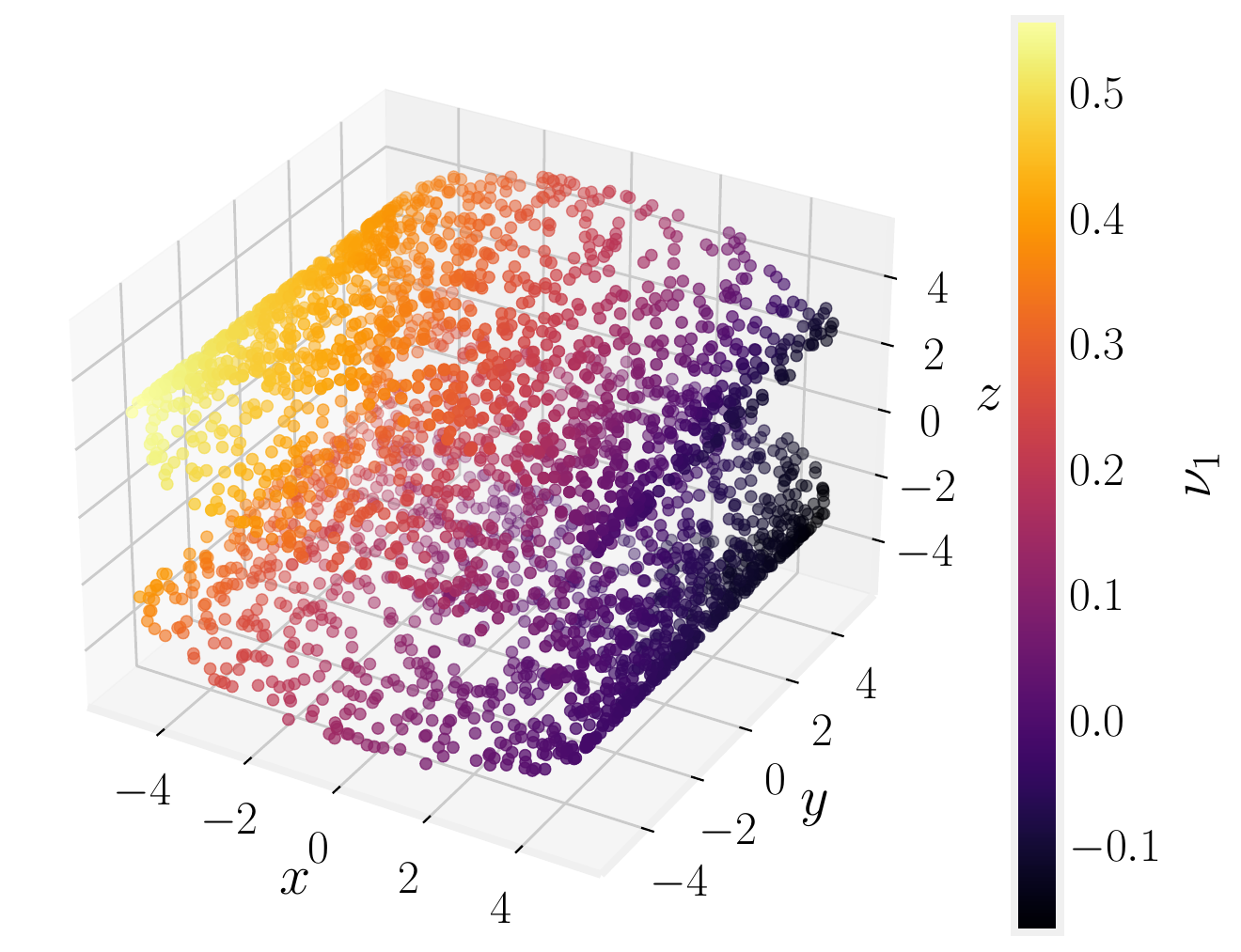}
			\caption{}
			\label{fig:ex_3_s_curve_nu_1}
		\end{subfigure}
		\begin{subfigure}[b]{0.33\textwidth}\includegraphics[width=\textwidth]{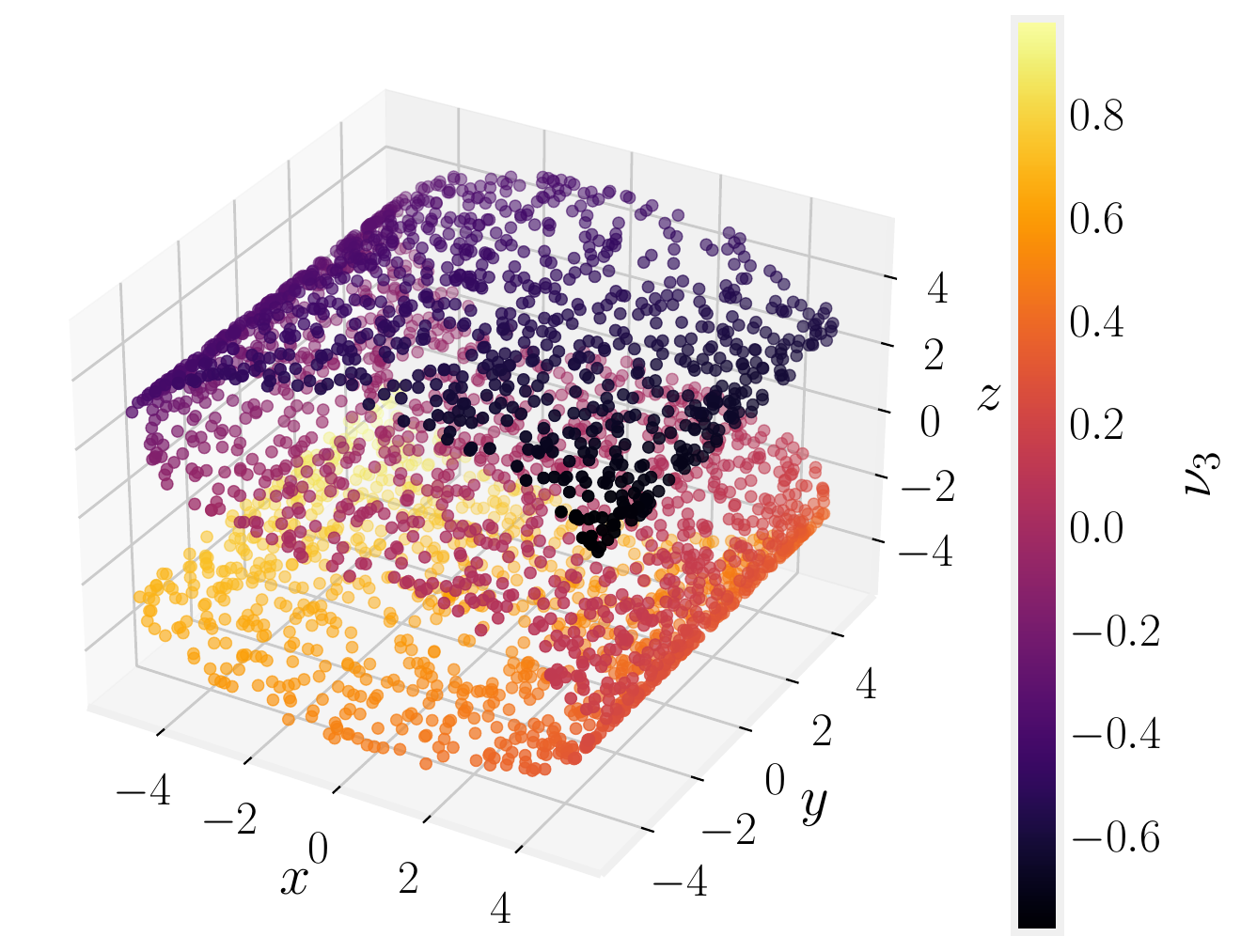}
			\caption{}
			\label{fig:ex_3_s_curve_nu_3}
		\end{subfigure}
		\caption{Optimization results for the S-curve data set from a single run of \cref{alg:CAE}. \cref{fig:ex_3_s_curve_latent} depicts the inferred two-dimensional representation of the training data, colored by the true arc-length in the ``long'' direction on the manifold. In \cref{fig:ex_3_s_curve_nu_1,fig:ex_3_s_curve_nu_3} the training set in ambient space is colored by the inferred latent coordinates ($\nu_1,\nu_3$).}
		\label{fig:ex_3_S_curve_embedding}
	\end{figure*}

It is instructive to look at the intermediate embeddings produced by the algorithm during training, which demonstrate how the approximating becomes sequentially more-compex and higher dimensional. This is demonstrated in figure \cref{fig:learning_transitions}.
	The S-Curve and Swiss Roll are standard test data sets for non-linear dimension reduction algorithms, being 2-dimensional but embedded in $\R^3$. Both feature large extrinsic curvature (which poses problems to the CAE training due to the spectral bias that accompanies neural networks) 
    and large Lipschitz constants of the embedding map. \cref{alg:CAE} is less stable, and does not always produce a 2-dimensional chart, but it is capable of doing so given a sufficiently good initialization.
	
	 \cref{fig:ex_3_S_curve_embedding} shows a `successful' embedding produced by the algorithm to the S-curve data set. The `natural' parametrization for the surface is a rectangle in $\R^2$ formed by the $y$-coordinate projection along with the arc length $l$ which forms the `s' shape when embedded in $\R^3$. Interestingly, while the CAE latent representation (\cref{fig:ex_3_s_curve_latent}) is two-dimensional, it is still non-linear, and retains some curvature properties of the original 3-dimensional embedding. Empirically, the shape of the latent space representation may depend on the choice of activation function for the autoencoder. For the particular embedding training error on $N=3000$ points is $\mcL_\text{CAE}=\num{1.7e-2}$ while the test error on $N=10000$ points is $L^2_\text{test}=1.6e-2$, obtained when setting the latent parameter equal to its mean during training $\nu_2^\text{test}=\E\qty{\nu_2^\text{train}}$.
\end{exmp}

\subsection{Training for PDE Examples}

\begin{figure}[h!]
        \begin{subfigure}[b]{0.54\textwidth}\includegraphics[width=\textwidth]{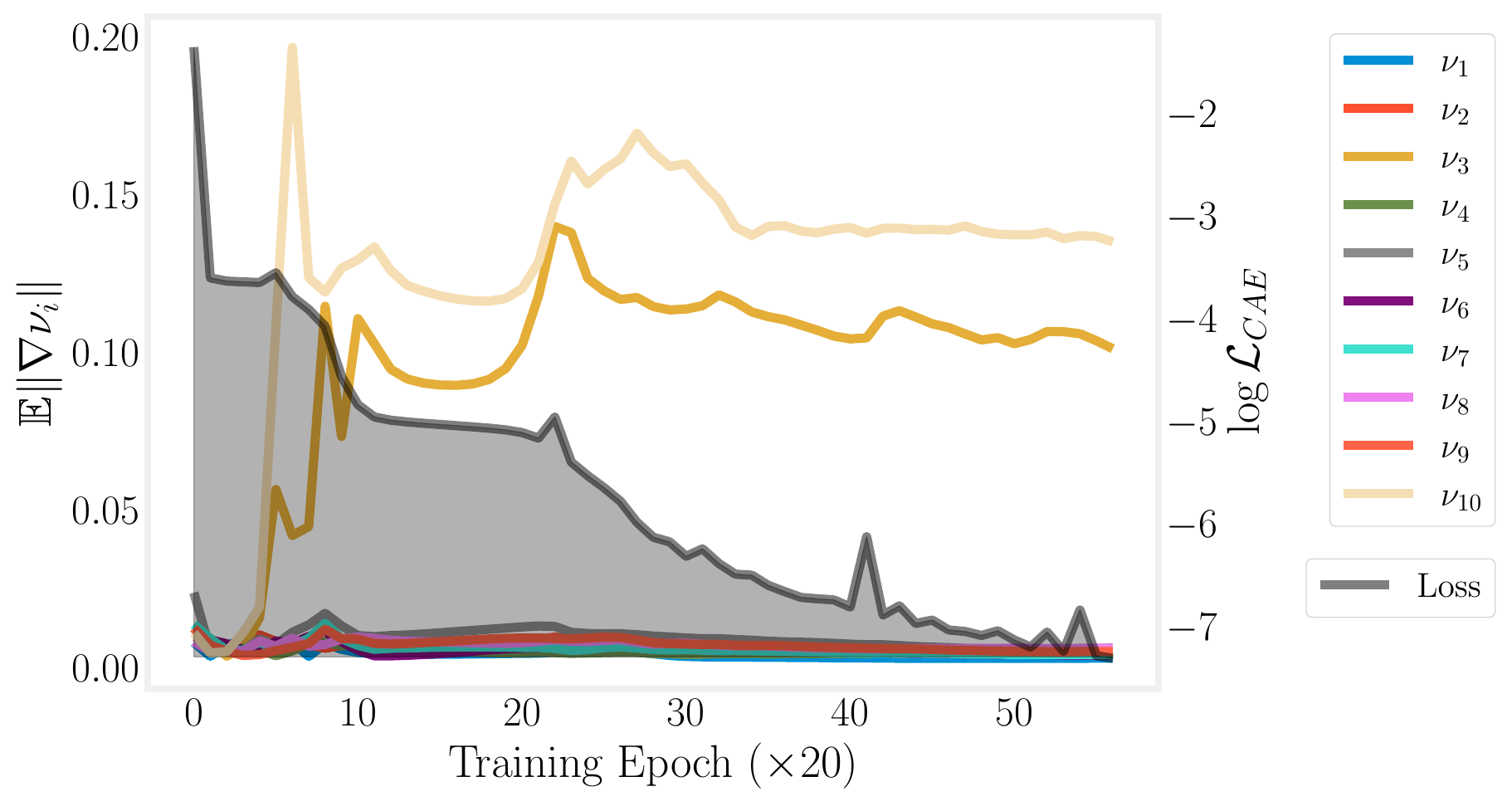}
		\caption{}
		\label{fig:ex_4_CI_gradient_loss}
	\end{subfigure}
	\begin{subfigure}[b]{0.45\textwidth}
		\includegraphics[width=\textwidth]{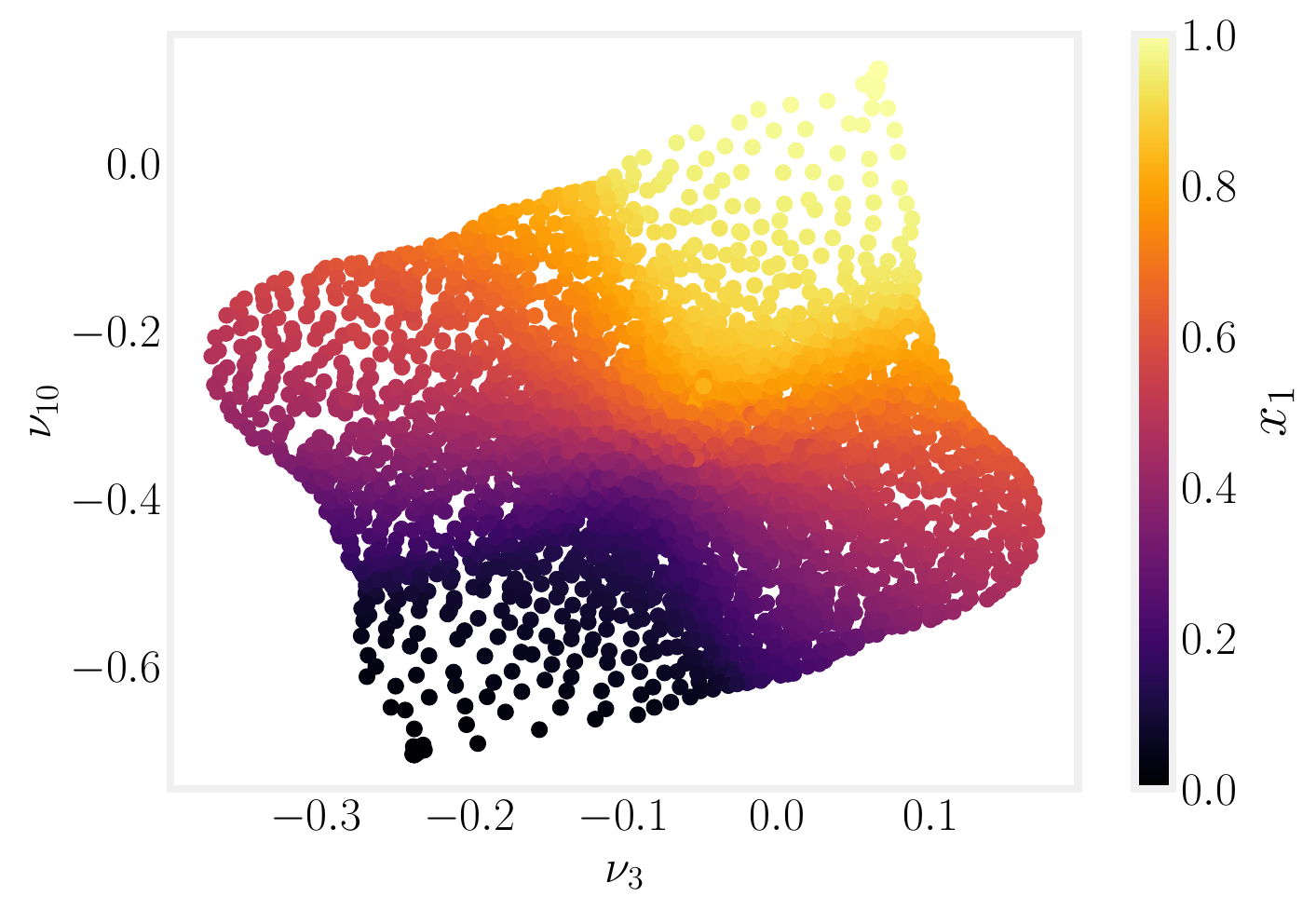}
		\caption{}
		\label{fig:PDE_embeddings_CI}
	\end{subfigure}
	\caption{(a) Training trajectory of \cref{alg:CAE} on the CI data set. (b) Data-driven embedding from \cref{exmp:KS} colored by one of the ambient coordinates of the given data set. }
	\label{fig:CI}
\end{figure}

\begin{figure*}[h!]
	\begin{subfigure}[b]{0.49\textwidth}
		\includegraphics[width=\textwidth]{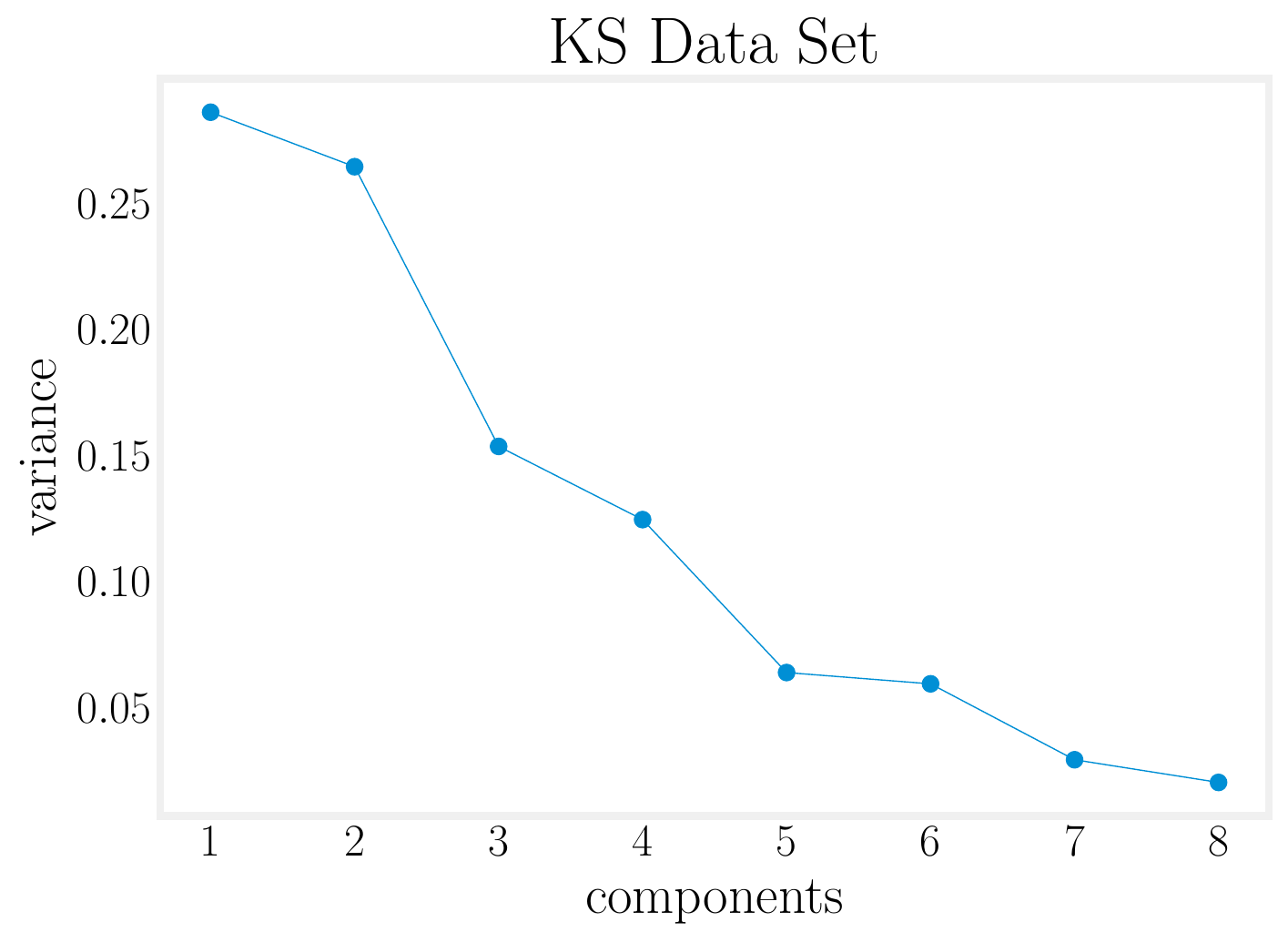}
		\caption{}
		\label{fig:ex_4_KS_pca}
	\end{subfigure}
	\begin{subfigure}[b]{0.49\textwidth}\includegraphics[width=\textwidth]{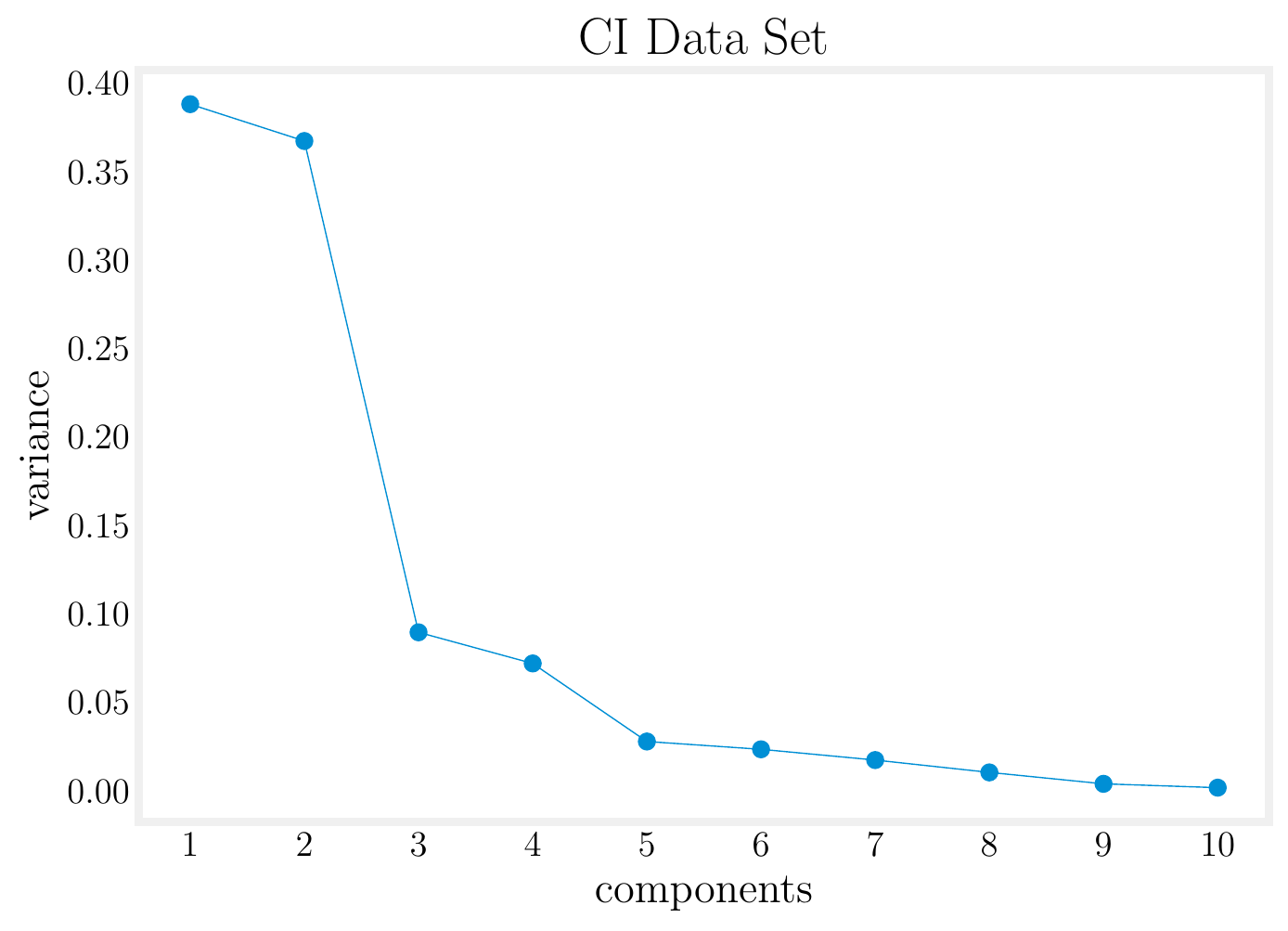}
		\caption{}
		\label{fig:ex_4_CI_pca}
	\end{subfigure}
	\caption{Explained variance of the corresponding principal components fit on the entire data set for \cref{exmp:KS} (KS - left) and \cref{exmp:CI} (CI - right)}
	\label{fig:ex_4_PCA}
\end{figure*}

\cref{fig:ex_4_PCA} shows the explained variance for a PCA fit on the PDE data sets. In each case, the "true" dimension of the non-linear submanifold is not immediately evident from the ratios of the variances. In particular, this suggests that a truncated linear approximation to each data set would not perform as well as the autoencoder neural networks.

\subsection{Bobbleheads}
\label{ap:bobbleheads}
We consider the bobblehead data set of \cite{lederman2014common} of grayscale (single color channel) bobblehead images $I_{t_i}$ observed at different times $t_i$. The image size is 320x250 pixels. Note that the original data set consists of coupled sets of images, i.e. observations of three bobbleheads by two cameras, in a disentanglement context. Here we \textit{only} consider one set of images of two rotating figures.

\begin{figure}
    \centering
    \begin{subfigure}{0.49\textwidth}
        \includegraphics[width=\textwidth]{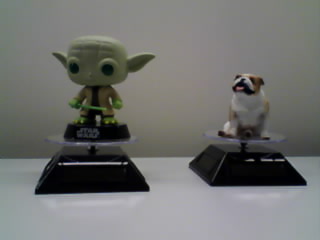}
		\caption{}
		\label{fig:bobblehead_setup}
	\end{subfigure}
        \begin{subfigure}{0.49\textwidth}
        \includegraphics[width=\textwidth]{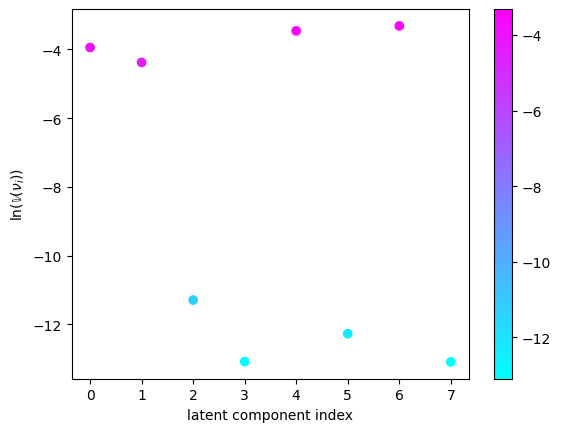}
		\caption{}
		\label{fig:bobblehead_variance}
	\end{subfigure}
    \caption{\textbf{(a)} Sample image of the bobblehead image data set featuring a Yoda and bulldog figure, which rotate at fixed (unknown) frequency, \textbf{(b)} Log Variance of the 8 latent components of the trained CAE. We observe $\nu_0,\nu_1,\nu_4,\nu_6$ to be active. The latent plot of \cref{fig:bobblehead_latent} shows the latent of the 60-dimensional data set for these selected components, revealing a toroidal structure in the latent space}
    \label{fig:ap_bobbleheads}
\end{figure}

We perform PCA on the $I_{t_i}$ data set and truncate to the first $60$ principal components, to originally compress the images. We then use a CAE with $8$ latent nodes and introduce the orthogonality constraint (as in \cref{alg:CAE}). The architecture of each encoder and decoder consists of 5 Tanh-activated layers followed by two linear layers, all of width 20. Upon convergence, we observe that the autoencoder identifies 4 latent dimensions, as judged by the latent variables with non-zero variance in \cref{fig:bobblehead_variance}.

After convergence, we fix the encoder weights (and therefore latent representation) and further train a larger decoder network (same layer structure with width 200) to achieve better reconstruction of the PCA components, which we can use to generate images by varying parameters in the latent representation.

\subsection{Concerning Approximations}
\label{sec:concerning_approx}
In the main theoretical result \cref{thm:orthogonal_charts} we assume the existence of a smooth conformal embedding $\Phi$ (and consequently a smooth chart), while in applications we may want to assume only, say, a $C^1$ chart. Common activation functions used in network architectures, e.g. the hyperbolic tangent or the sigmoid function, are smooth. In this case, we note that by the Meyers-Serrin Theorem (\cite{evans2022partial}, Section 5.3.2, Theorem 2), $C^\infty(\mathcal{\mcM})\cap W^{k,p}(\mathcal{\mcM})$ is dense in $W^{k,p}(\mathcal{\mcM})$ for $1\leq p<\infty$, and furthermore, sufficiently large networks are dense in $C^\infty(\mathcal{\mcM})\cap W^{k,p}(\mathcal{U})$ (\cite{czarnecki2017sobolevNIPS,HORNIK1990551}). Thus, as long as orthogonal charts on $\mathcal{\mcM}$ exist, the described CAE architectures are sufficiently expressive to approximate these maps properly, for sufficiently large network sizes. Of course, the proposed loss function (\cref{eqn:CAE_loss}) is generically non-convex, so that gradient descent algorithms are not guaranteed to converge to a global minimizer. Additionally, the orthogonality term is not strictly minimized: upon convergence it is only satisfied within a small error tolerance. This can become problematic in practice, since gradients with sufficiently small norm may also appear to satisfy the constraint in that manner. A more subtle additional point is the following: a $C^\infty$ network is not only $C^\infty$ on the data domain ($\mathcal{\mcM}$) but also on the entire embedding space $\R^n$. This space is very regular! For example, the method will fail to embed a circle ($S^1\subset\R^2$), since any smooth chart for $S^1$ must have a singularity in its interior. Other issues emerge when data sets have small Euclidean distances coupled with large geodesic distances, since a network may erroneously ``connect" such points (\cref{exmp:circle,exmp:S_curve}). These issues are not specific to our class of autoencoders nor to the functional we are minimizing.

\clearpage

\begin{figure*}
    \centering
    \includegraphics[width=\linewidth]{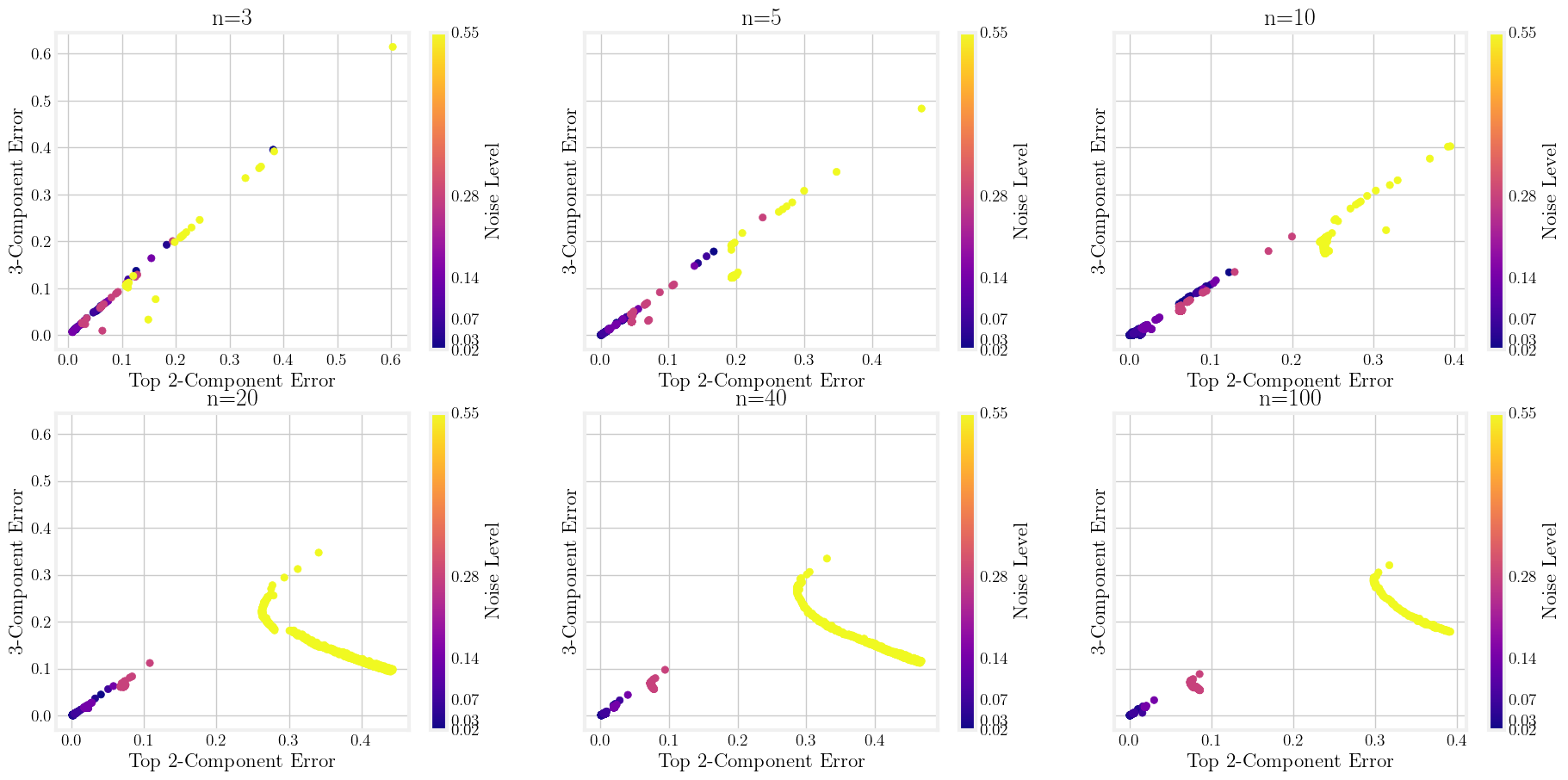}
    \caption{Comparison between the 3-component reconstruction (using the full available latent space) vs. the Top 2-component reconstruction. Each plot corresponds to a single fixed embedding dimension $n=\qty{3,5,10,20,40,100}$, and each color to added ambient Gaussian noise with a different choice of standard deviation $\sigma=\qty{0.02, 0.03, 0.07, 0.14, 0.28,
       0.55}$.}
    \label{fig:robustness_fixed_dim}
\end{figure*}

\begin{figure*}
\centering
	\begin{subfigure}[b]{0.45\textwidth}
		\includegraphics[width=\textwidth]{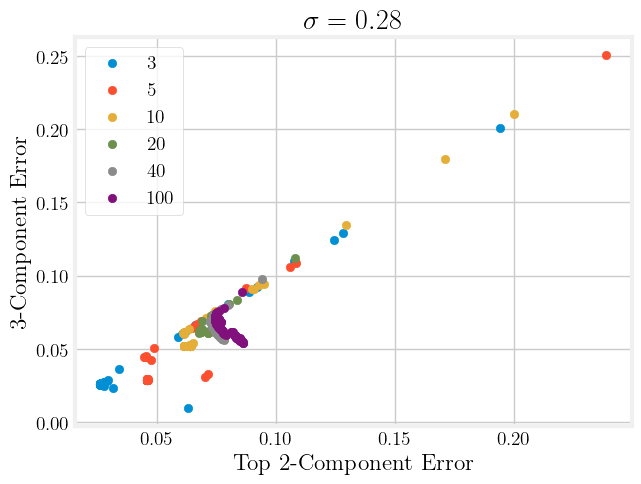}
		\caption*{}
	\end{subfigure}
	\begin{subfigure}[b]{0.45\textwidth}\includegraphics[width=\textwidth]{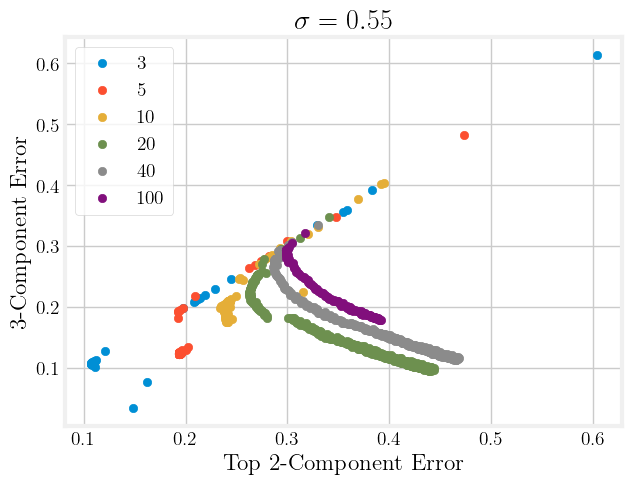}
		\caption*{}
	\end{subfigure}
	\caption{Comparison between 3-component reconstruction (using the full available latent space) vs. the top 2-component reconstruction. Each plot corresponds to a fixed ambiant noise level (Gaussian with standard deviation $\sigma=\qty{0.28,0.55}$), and each color corresponds to a different dimension, listed in the legend.}
	\label{fig:robustness_fixed_noise}
\end{figure*}

\section{Robustness Studies}\label{ap:robustness}

We are interested in characterizing the robustness of the proposed methodology, in particular \cref{alg:CAE}, when the ambient dimension $n$ and amount of noise in the training data varies. To this end, we devise the following experiment.

The training data of \cref{exmp:toy} (which is two-dimensional embedded in $k=3$-dimensional Euclidean space) is embedded in Euclidean space of increasing dimension $n=\qty{5,10,20,40,100}$ using a random $n\times k$ truncated unitary matrix. Then, normal ambient-space noise is added with standard deviation $\sigma=ld$ where $d=\sqrt{3}$ is the approximate diameter of the data, and $l=\qty{0.01, 0.02, 0.04, 0.08, 0.16, 0.32}$

For each combination of $n$ and $\sigma$, we train an AE architecture with three latent nodes to reconstruct the prescribed high-dimensional noisy data. It is clear that with sufficient noise, the `intrinsic' low-dimensional structure will inevitably be lost. In \cref{fig:robustness_fixed_dim} we plot the reconstruction error achieved with all three latent components ($y$-axis) compared to the reconstruction error achieved with the top-2 ($x$-axis, where `top' is measured by the average $\ell^2$-norm of the gradient over the training data). When lying on the diagonal, these errors are approximately equal, signifying that the `correct' dimension is inferred. That is, the training algorithm is not making use of the third available component to achieve a `good' reconstruction of the samples. However, when the samples stray towards the lower half, it signifies that the algorithm has over-estimated the dimension, and the third latent component is used.

We observe that for small levels of noise, the architecture is able to identify the `correct dimension' to a good level of accuracy, as demonstrated by the darker colors in \cref{fig:robustness_fixed_dim}. However, after a certain level of accuracy, the lighter colors (especially pink and yellow) stray off the diagonal, indicating a misidentification of the dimension. It is further interesting to visualize the behavior of the algorithm for the same amount across increasing dimension. This is demonstrated in \cref{fig:robustness_fixed_noise}, where it is possible to see that the accuracy by which we can approximate the embedded data with two components decreases with dimension. We do not study the rate at which this phenomenon occurs here, since it may be data and architecture dependent.

\textbf{Procedural Details}
To perform the experiments that result in \cref{fig:robustness_fixed_dim,fig:robustness_fixed_noise}, we initialize an architecture in which the encoder and decoder have the same size, with 5 fully-connected $\tanh$ layers followed by 2 linear layers. The width  $w$ of each layer is increasing with ambient dimension as:
\begin{align*}
    w=\mathrm{int}(10\ceil{\sqrt{n}})
\end{align*}
This is done because we empirically find that larger networks are needed to get similar level of accuracy in higher dimensions.

We stop training the architecture if the loss plateaus for sufficiently long (1500 epochs), or if the $\ell^2$-reconstruction loss $\mathcal{L}$ satisfies:
\begin{align*}
    \mathcal{L}\leq\max\qty{\num{5e-4},\frac{\sigma^2\sqrt{n/3}}{10}}
\end{align*}
reflecting the idea that the threshold should be increasing in ambient-space dimension. We observe that for higher noise levels and higher dimensions, the loss reaches a plateau before the threshold accuracy is achieved, since the latent space is always 3-dimensional and therefore cannot well-approximate the noisy high-dimensional object.

\clearpage

\section{Benchmarking Specifics}
\label{app:benchmark_specs}
We describe the mathematical formulation of each autoencoder variant benchmarked in \cref{tab:ae-benchmark}.

\subsection{Simple Autoencoder}
The standard autoencoder \cite{kramer1991nonlinear} consists of an encoder $f_\theta : \mathbb{R}^d \to \mathbb{R}^k$ and a decoder $g_\phi : \mathbb{R}^k \to \mathbb{R}^d$ trained to minimize the mean squared reconstruction error:
\begin{equation}
    \mathcal{L}_{\text{AE}} = \frac{1}{n} \sum_{i=1}^n \left\| x_i - g_\phi(f_\theta(x_i)) \right\|^2.
\end{equation}
No regularization is applied to the latent space. The parameters $\theta$ and $\phi$ denote the space of weights and biases used to specify each network. In our experiments, the networks are fully connected feed-forward MLPs.

\subsection{Sparse Autoencoder with \texorpdfstring{$L^1$}{L1} Regularization}
The $L^1$-regularized autoencoder (see e.g. \cite{empirical_sparse})  adds a penalty on the absolute value of the latent activations to encourage sparsity:
\begin{equation}
    \mathcal{L}_{\text{Sparse-}L^1} = \frac{1}{n} \sum_{i=1}^n \left\| x_i - g_\phi(f_\theta(x_i)) \right\|^2 + \alpha \cdot \frac{1}{n} \sum_{i=1}^n \left\| f_\theta(x_i) \right\|_1,
\end{equation}
where $\alpha > 0$ is a regularization coefficient.

\subsection{Sparse Autoencoder with KL Divergence}
This model encourages the average activation of each latent unit to match a target sparsity $\rho \in (0,1)$ using a Kullback-Leibler divergence penalty (see e.g. \cite{empirical_sparse}). Let $\hat{\rho}_j = \frac{1}{n} \sum_{i=1}^n z_{ij}$ denote the mean activation of the $j$th latent unit across the dataset. The KL penalty is
\begin{equation}
    \mathrm{KL}(\rho \| \hat{\rho}_j) = \rho \log \frac{\rho}{\hat{\rho}_j} + (1 - \rho) \log \frac{1 - \rho}{1 - \hat{\rho}_j}.
\end{equation}
The total loss is
\begin{equation}
    \mathcal{L}_{\text{Sparse-KL}} = \frac{1}{n} \sum_{i=1}^n \left\| x_i - g_\phi(f_\theta(x_i)) \right\|^2 + \alpha \cdot \sum_{j=1}^k \mathrm{KL}(\rho \| \hat{\rho}_j).
\end{equation}

\subsection{Gradient Norm Autoencoder}
This model penalizes the squared norm of the input gradient of each latent variable:
\begin{equation}
    \mathcal{L}_{\text{GradNorm}} = \frac{1}{n} \sum_{i=1}^n \left\| x_i - g_\phi(f_\theta(x_i)) \right\|^2 + \alpha \cdot \sum_{j=1}^k \left\| \nabla_x \nu_j(x_i) \right\|^2,
\end{equation}
where $\nu_j(x) = f_{\theta,j}(x)$ is the $j$th latent component. This regularization encourages each latent to vary smoothly and independently with respect to the input.

\subsection{Beta-Variational Autoencoder (Beta-VAE)}
The $\beta$-VAE (\cite{higgins2017beta}) augments the standard VAE (\cite{kingma2013vae}) objective by scaling the KL divergence term:
\begin{equation}
    \mathcal{L}_{\beta\text{-VAE}} = \mathbb{E}_{q_\theta(z|x)}\left[ \| x - g_\phi(z) \|^2 \right] + \beta \cdot D_{\text{KL}} \left( q_\theta(z|x) \| p(z) \right),
\end{equation}
where $q_\theta(z|x)$ is the encoder distribution, $p(z)$ is a standard Gaussian prior, and $\beta > 0$ controls the strength of regularization. A large $\beta$ encourages more disentangled representations at the expense of reconstruction fidelity.
\clearpage
\subsection{Extended Results}
\begin{table}[ht]
\centering
\caption{Estimated intrinsic dimension and reconstruction error across autoencoder models trained on the Chafee-Infante (CI) and Kuramoto-Sivashinsky (KS) PDE datasets. Each model with a given parameter selection was run five times and the mean and standard deviation of the reconstruction error are presented in the table along with the mode of the predicted instrinsic dimension.}
\label{tab:merged-ae-results}
\begin{tabular}{|l|l|cc|cc|}
\hline
\multirow{2}{*}{\textbf{Model}} & \multirow{2}{*}{\textbf{Param(s)}} & \multicolumn{2}{c|}{\textbf{CI PDE (ID = 2)}} & \multicolumn{2}{c|}{\textbf{KS PDE (ID = 3)}} \\
\cline{3-6}
& & \textbf{Est. ID} & \textbf{Error ($\pm$ std)} & \textbf{Est. ID} & \textbf{Error ($\pm$ std)} \\
\hline
CAE & $\alpha = 0.01$ & 10 & $2.0\text{e-}5 \pm 1\text{e-}6$ & 6 & $1.6\text{e-}5 \pm 3\text{e-}6$ \\
& $\alpha = 0.1$ & 6 & $3.0\text{e-}5 \pm 2\text{e-}6$ & 4 & $6.0\text{e-}5 \pm 2\text{e-}6$ \\
& $\bm{\alpha = 1.0}$ & \textbf{2} & $\bm{3.6\text{e-}5 \pm 1.5\text{e-}6}$ & \textbf{3} & $\bm{4.4\text{e-}5 \pm 8\text{e-}6}$ \\
& $\alpha = 10.0$ & 3 & $7.4\text{e-}5 \pm 1.2\text{e-}6$ & 3 & $4.6\text{e-}5 \pm 1.1\text{e-}5$ \\
\hline
Sparse AE ($L^1$) & $\alpha = 0.001$ & 10 & $1.5\text{e-}4 \pm 3\text{e-}5$ & 6 & $6.8\text{e-}5 \pm 2.3\text{e-}6$ \\
& $\alpha = 0.01$ & 3 & $2.31\text{e-}4 \pm 2\text{e-}5$ & 4 & $3.7\text{e-}4 \pm 2.5\text{e-}5$ \\
& $\alpha = 0.1$ & 2 & $5.7\text{e-}4 \pm 2\text{e-}5$ & 3 & $4.2\text{e-}4 \pm 1.7\text{e-}5$ \\
& $\alpha = 1.0$ & 2 & $5.5\text{e-}3 \pm 2.5\text{e-}4$ & 2 & $4.1\text{e-}3 \pm 1.2\text{e-}3$ \\
\hline
Sparse AE (KL) & $\rho{=}0.01,\alpha{=}0.001$ & 10 & $1.82\text{e-}4 \pm 2\text{e-}5$ & 4 & $2.2\text{e-}4 \pm 2\text{e-}5$ \\
& $\rho{=}0.01,\alpha{=}0.01$ & 10 & $2.4\text{e-}4 \pm 1\text{e-}5$ & 6 & $3.2\text{e-}4 \pm 1.5\text{e-}5$ \\
& $\rho{=}0.01,\alpha{=}0.1$ & 10 & $1.5\text{e-}4 \pm 1.2\text{e-}5$ & 5 & $3.0\text{e-}4 \pm 1.2\text{e-}5$ \\
& $\rho{=}0.01,\alpha{=}1.0$ & 10 & $2.4\text{e-}4 \pm 1.1\text{e-}5$ & 8 & $1.8\text{e-}3 \pm 1.4\text{e-}4$ \\
\hline
Grad. Norm AE & $\alpha = 0.001$ & 10 & $2.0\text{e-}5 \pm 1\text{e-}5$ & 8 & $7.1\text{e-}5 \pm 2\text{e-}5$ \\
& $\alpha = 0.01$ & 9 & $2.3\text{e-}4 \pm 1.5\text{e-}5$ & 7 & $6.3\text{e-}5 \pm 1.8\text{e-}5$ \\
& $\alpha = 0.1$ & 5 & $4.7\text{e-}3 \pm 3.6\text{e-}4$ & 5 & $3.8\text{e-}4 \pm 1.6\text{e-}5$ \\
& $\alpha = 1.0$ & 3 & $3.0\text{e-}3 \pm 2.0\text{e-}4$ & 6 & $3.2\text{e-}3 \pm 1.2\text{e-}4$ \\
\hline
$\beta$-VAE & $\beta = 10^{-6}$ & 9 & $3.2\text{e-}5 \pm 0.5\text{e-}6$ & 8 & $2.2\text{e-}5 \pm 3\text{e-}6$ \\
& $\beta = 10^{-5}$ & 10 & $4.0\text{e-}5 \pm 0.7\text{e-}6$ & 8 & $5.0\text{e-}5 \pm 2\text{e-}6$ \\
& $\beta = 10^{-4}$ & 10 & $1.3\text{e-}4 \pm 1.2\text{e-}5$ & 8 & $1.7\text{e-}4 \pm 1.1\text{e-}5$ \\
& $\beta = 10^{-3}$ & 10 & $1.1\text{e-}3 \pm 1.5\text{e-}4$ & 8 & $2.9\text{e-}3 \pm 1.4\text{e-}4$ \\
\hline
Simple AE & -- & 10 & $4.8\text{e-}5 \pm 6\text{e-}6$ & 8 & $3.5\text{e-}5 \pm 3\text{e-}6$ \\
\hline
\end{tabular}
\end{table}

In \cref{tab:merged-ae-results} we compare model performance on the PDE data sets of \cref{sec:Numerical_Examples} for different values of model parameters, i.e. weights of the regularizing terms in the corresponding loss defined above. Overall, while all models are sensitive to the choice of these parameters, CAEs and $L^1$-regularized sparse autoencoders are empirically observed to have better performance at estimating the true intrinsic dimension of the data set, while between the two, CAEs achieve better reconstruction error on average. Once again, the intrinsic dimension is computed by succesively performing the data reconstruction with the top-$k$ latent components ranked by variance, and determining the minimal $k$ for which the reconstruction error is approximately equal to the total reconstruction error. The latent space for \textit{all} architectures is 10-dimensional for the CI example and 8-dimensional for the KS example.

\section{Orthogonal Charts}
\label{app:Diff_Geo}

For positive integers $n, k\in\N$ with $k\leq n$, let $\mathcal{M}\subset\R^n$ be a $k$-dimensional smooth manifold, and $\mathcal{U}\subseteq\mathcal{M}$ be a precompact, simply connected, open subset equipped with a single chart $\psi:\mathcal{U}\to\R^k$. Note that $\psi$ defines a diffeomorphism between $\mathcal{U}$ and $\mathcal{V}\doteq \psi(\mathcal{U})$. Furthermore, let $\vb{x}=(x_1,...,x_n)\in\mathcal{U}$ and define $\psi$ component-wise as
\begin{align}
	\psi(\vb{x})=\mqty(\psi_1(\vb{x})\\ \psi_2(\vb{x})\\ \vdots\\\psi_k(\vb{x}))
\end{align}
with each $\psi_i\in C^\infty(\mathcal{U},\R)$.

In this setting, at any point $p\in\mathcal{M}$, the tangent space $T_p\mathcal{M}$ is a copy of $\R^k$ and is spanned by $k$ linearly independent vectors. One way of obtaining a frame for this vector space is by considering the set of gradients of the components of $\psi$, $E_p$. Subsequently, one may proceed to orthogonalize (or orthonormalize) on $E_p$ through the GS algorithm. Note that at every $p\in\mathcal{U}$, each vector in $E_p$ must be linearly independent since otherwise $\psi$ could not be a diffeomorphism. Denote this set by $E_p^\perp$ and observe that $\abs{E_p^\perp}=\abs{E_p}=k$.

\begin{gather}
	E_p = \qty{\grad_p^\mcM\psi_i:i\leq k}\\
	E_p^\perp = \text{GS}(E_p)
\end{gather}

While this method yields an orthogonal frame on $T\mcM$, it is not always possible to find any \textit{coordinates} that produce such orthogonal frames on manifolds (which is equivalent to the manifold being \textit{conformally flat}: a distribution is not guaranteed to be involutive even if it is pointwise orthogonal, and hence not guaranteed to be integrable through Frobenius's theorem). All two-dimensional surfaces are conformally flat (and in three dimensions there exist local orthogonal coordinates even if a manifold is \textit{not} conformally flat (i.e. the Cotton tensor does not vanish \cite{johnson2023orthogonal}). In four dimensions and above the property is equivalent to the vanishing of the Weyl tensor \cite{DeTurckDennisM.1984Eoed}. One might hope that by controlling the eigenvalues of the Weyl tensor we may be able to infer properties of a manifold \cite{onti2017almost_conformally_flat}. The exact behavior of (not only our) proposed algorithms in such a case warrants further study.

\textbf{Proof of \cref{thm:orthogonal_charts}}
Because $f$ is already a diffeomorphism on $\mcM$ (with it's image), but has higher dimension $f:\R^n\to\R^n$, the only non-trivial part is to show that the \textit{same} components of $f$ are non-constant on the entirety of $\mcM$. I.e. there is no component which is constant on \textit{part}, but not the whole of $\mcM$. We do so below:

It is simple to first argue for the case where $\mcM$ is a (intrinsically 1-dimensional) curve embedded in $\R^2$. We then extend to higher dimensions. We provide a topological proof, though one can also follow a more constructive geometric direction.

Note that for any smooth vector field $X\in\mathfrak{X}(\R^n)$ can be smoothly projected onto the tangent space of a smooth submanifold $T_p\mcM$ by making use of an arbitrary frame on an open neighborhood $U$, and Gram-Schmidt (\cite{Lee00}, Chapter 8).

Now, suppose that $f$ is a diffeomorphism on $\mcM$ and $\mcL f=0$. The number of non-vanishing components $\qty{\grad^\mcM f_i}_{i=1}^n$ must be at most $k$, since $T_p\mcM$ is $k$-dimensional, and we can find at most $k$ linearly independent vectors spanning it. Furthermore, since $f$ is a diffeomorphism (on $\mcM$), there cannot be a point with ferwer than $k$ such vectors, since otherwise the Jacobian of $f$ would be singular on $\mcM$ and $f$ could not be a diffeomorphism.

In 2 dimensions: Let $f=(f_1,f_2)$ and let $U,V$ be the sets (subsets of $\mcM$) on which $\grad^\mcM f_1,\grad^\mcM f_2$ respectively vanish. Due to the smoothness of the vector field projection, for every point $u\in U$ (resp. $v\in V$) there exists an open neighborhood centered at $u$ also in $U$ (resp. centered at $v$ in $V$) and so $U$ and $V$ are both open sets. Because of the diffeomorphism constraint, we must have $U\cap V=\emptyset$. We also have $\mcM=U\cup V$, and since $U\cap V=\emptyset$ we also have that $V$ is the complement of $U$, and must therefore be closed. Thus, $V$ is both open and closed and must either be the empty set, or $\mcM$.

In $n$ dimensions: Let $f=(f_1,...,f_n)$, and for a point $u\in\mcM$, pick the components of $f$ whose projected gradient does not vanish on $\mcM$, denoted by $f\vert_k$, $k\in K$ being the appropriate index set over $[1,...,n]$. Let $U$ be the set of points where all of $\grad^\mcM f\vert_k$ do not vanish, and $V$ be the set where at least one of the $\grad^\mcM f\vert_k$ vanishes (at any given point there must be another index set $K'$ of cardinality $k$, for which the gradients do not vanish. We denote this by $f\vert_{k'}$). That is:
\begin{align*}
    U&=\qty{p\in\mcM:\grad^\mcM_pf_k\neq 0,k\in K}\\
    V&=\qty{p\in\mcM:\grad^\mcM_pf_k=0\qq{for some}k\in K}
\end{align*}
 We argue that $U$ and $V$ are open: For every point in $u\in U$ there is a neighborhood the point where, by smoothness of the projected vector fields, the gradients of $f\vert_k$ do not vanish. Similarly, for every point $v\in V$, there is a neighborhood around the point where the gradients of $f\vert_{k'}$ do not vanish, from which it follows that if a point has $\grad^\mcM f_i=0$ for at least \textit{one} component of $f\vert_k$, there is an open neighborhood around it which also satisfies $\grad^\mcM f_i=0$ for the same component (due to the orthogonality constraint). Now, clearly $U\cup V=\mcM$, and additionally, $U\cap V=\emptyset$. Since $V=U^c$, V is both open and closed, and so must be empty, or $\mcM$.


\end{document}